\documentclass[11pt]{article}

\usepackage[preprint]{acl}

\usepackage{times}
\usepackage{latexsym}

\usepackage[T1]{fontenc}

\usepackage[utf8]{inputenc}

\usepackage{microtype}

\usepackage{inconsolata}

\usepackage{graphicx}

\usepackage{subcaption}
\usepackage{makecell}
\usepackage{booktabs}
\usepackage{threeparttable}
\usepackage{hyperref}

\newcommand{\systemName}{MTQE.en-he}

\title{\systemName{}: Machine Translation Quality Estimation for English-Hebrew}

\author{Andy Rosenbaum \\
  Independent Researcher \\
  New York, USA \\
  \texttt{me@andyjrosenbaum.com} \\\And
  Assaf Siani \\
  Lexicala \\
  Tel Aviv, Israel \\
  \texttt{assaf@lexicala.com} \\\And
  Ilan Kernerman \\
  Lexicala \\
  Tel Aviv, Israel \\
  \texttt{ilan@lexicala.com}}

\begin{document}
\maketitle
\begin{abstract}
We release \systemName{}: to our knowledge, the first publicly available English-Hebrew
benchmark for Machine Translation Quality Estimation.
\systemName{} contains 959 English segments from WMT24++, each paired with a machine translation into Hebrew,
and Direct Assessment scores of the translation quality
annotated by three human experts.
We benchmark ChatGPT prompting, TransQuest, and CometKiwi and show that ensembling
the three models outperforms the best single model (CometKiwi) by 6.4 percentage points Pearson and 5.6 percentage points Spearman.
Fine-tuning experiments with TransQuest and CometKiwi reveal that full-model updates are sensitive to overfitting and distribution collapse, yet parameter-efficient methods (LoRA, BitFit, and FTHead, i.e., fine-tuning only the classification head) train stably and yield improvements of 2-3 percentage points. \systemName{} and our experimental results enable future research on this under-resourced language pair.
\end{abstract}

\section{Introduction}

Machine Translation Quality Estimation (MTQE) is an important step in
Machine Translation (MT) pipelines: by scoring translations,
the overall system can accept high quality
outputs and flag low quality
outputs for human review or post-editing.

Hebrew is considered a mid-resource language in Natural Language Processing (NLP), with substantially fewer annotated resources than high-resource languages like English, and its rich morphology leads to effective low-resource learning conditions \citep{joshi-etal-2020-state,tsarfaty-etal-2019-whats}.

We create and release\footnote{\href{https://gitlab.com/lexicala-public/mtqe-en-he}{gitlab.com/lexicala-public/mtqe-en-he}} \systemName{}, a new dataset for English-Hebrew MTQE consisting 
of 959 segments annotated by three human experts and benchmark popular models (ChatGPT, \mbox{TransQuest}, and CometKiwi) on it,
showing strong baselines yet plenty of room for improvement.

\section{Related Work}

Quality Estimation (QE) has been studied for years as a
WMT shared task.
While WMT23 QE includes a test set from Hebrew to English,
generally translating from a high-resource language (English) into
a lower resource language (Hebrew) is a more difficult task.
\citet{Zhao_2024} provide an overview
of the history of MTQE.

TransQuest \citep{ranasinghe-etal-2020-transquest} fine-tunes XLM-RoBERTa-large
\citep{conneau-etal-2020-unsupervised} on Quality Estimation and explores cross-lingual transfer.
CometKiwi \citep{rei-etal-2021-references} simultaneously models and predicts sentence-level and word-level quality.
\citet{juraska-etal-2024-metricx} build MetricX-24, a model that can handle quality
estimation both with and without references, and employ synthetic data generation to improve the results.

The closest relative of our work is EPOQUE \citep{jafari-harandi-etal-2024-epoque},
who develop a similar test set for English-Persian, then benchmark and fine-tune TransQuest.
The novelty of our work is threefold (1) we are the first
to publicly release a dataset for English-Hebrew,
(2) we benchmark ChatGPT prompting and ensembling models,
and (3) we explore parameter-efficient fine-tuning of TransQuest and CometKiwi.

\section{Dataset}

\begin{figure*}[t]
  \centering
  \begin{subfigure}{0.49\linewidth}
    \includegraphics[width=\linewidth]{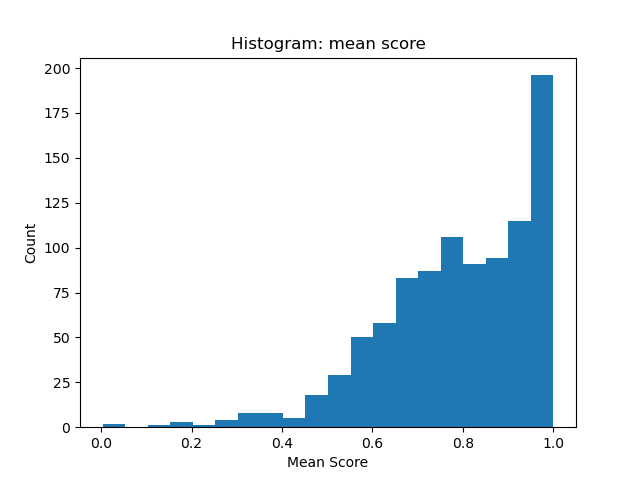}
    \caption{Mean score distribution.}
    \label{fig:allScoreDist}
  \end{subfigure}
  \hfill
  \begin{subfigure}{0.49\linewidth}
    \includegraphics[width=\linewidth]{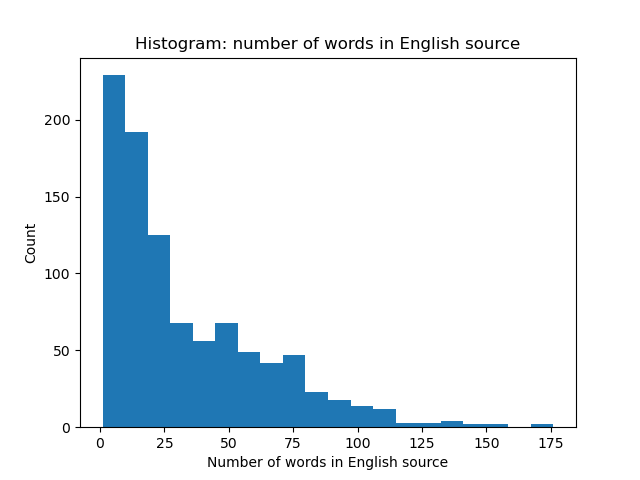}
    \caption{English source word length distribution.}
    \label{fig:allNumWordsDist}
  \end{subfigure}
  \caption{\systemName{} dataset statistics (n=959). Left: Mean Score. Right: Number of words in English source.}
  \label{fig:dataStats}
\end{figure*}

\systemName{} contains 959 English segments\footnote{The original dataset is 998 samples. We remove the 38 samples
flagged with \texttt{is\_bad\_source}, and one duplicate.}
 from English-Hebrew WMT24++ 
\citep{deutsch2025wmt24expandinglanguagecoverage}, covering four domains:
literary, news,
social, and speech.
We apply Google Translate on the English source
to get the Hebrew translation,
then hire three human experts to annotate the segments with a
standard Direct Assessment (DA) score
in the range of 0 to 100. (Guidelines in Appendix \ref{sec:chatGPTPrompts}.)

All three annotators have native-level
proficiency in English and Hebrew, plus
extensive linguistics background and experience with translation workflows.
The annotators are professional contacts of the paper authors and were paid fairly for their work in their locale.
The inter-annotator agreement (Pearson one vs. mean of others) is 0.5168, 0.5270, and 0.5573, respectively, for a mean of 0.5337,
which is reasonable for MTQE DA labels.

We show the histogram of mean score (left) and the number of words in the English source (right) in Figure \ref{fig:dataStats}.
We show the frequency of representative score ranges
in Table \ref{tab:scoreRangeCounts}.
The scores trend high (73\% scored 70 and above)
and segment lengths trend short
(59\% of the segments are under 30 words).

\begin{table}[h]
\centering
\begin{tabular}{lcc}
\toprule
 Mean score range   & Count     & Percentage   \\
\midrule
 $[91, 100]$         & 300 / 959 & 31\%         \\
 $[70, 91)$          & 402 / 959 & 42\%         \\
 $[51, 70)$           & 205 / 959 & 21\%         \\
 $[0, 51)$            & 52 / 959 & 6\%         \\
\bottomrule
\end{tabular}
\caption{Frequency of representative score ranges.}
\label{tab:scoreRangeCounts}
\end{table}

\section{Benchmarks on the Full Dataset}

\begin{figure*}[t]
  \centering
  \begin{subfigure}{0.32\linewidth}
    \includegraphics[width=\linewidth]{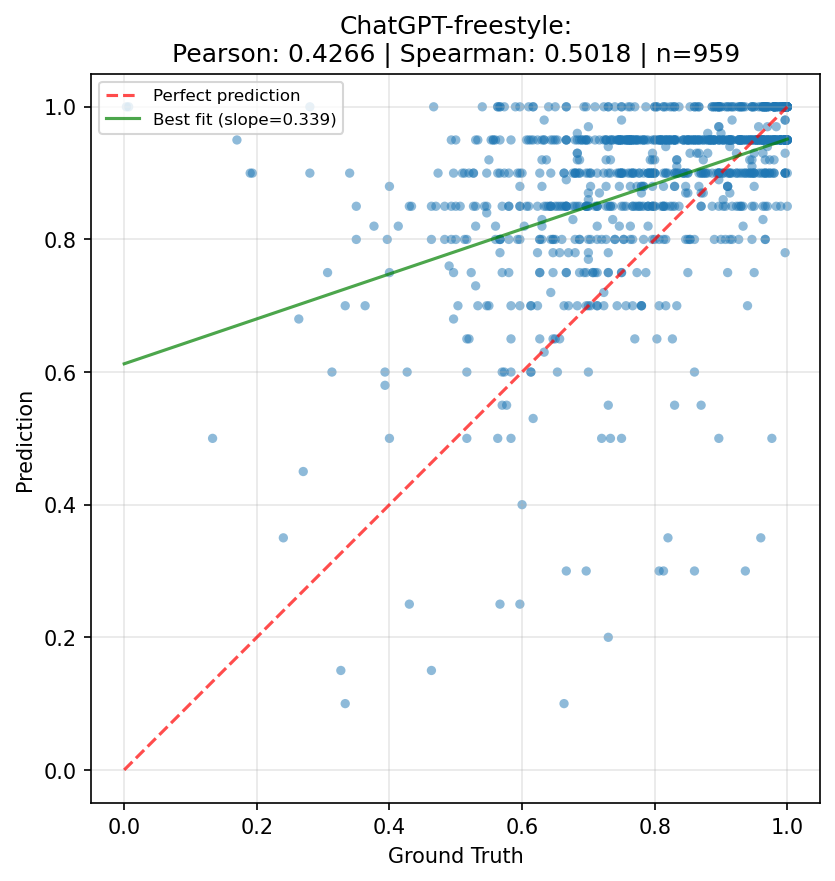}
    \caption{ChatGPT-freestyle}
  \end{subfigure}
  \hfill
  \begin{subfigure}{0.32\linewidth}
    \includegraphics[width=\linewidth]{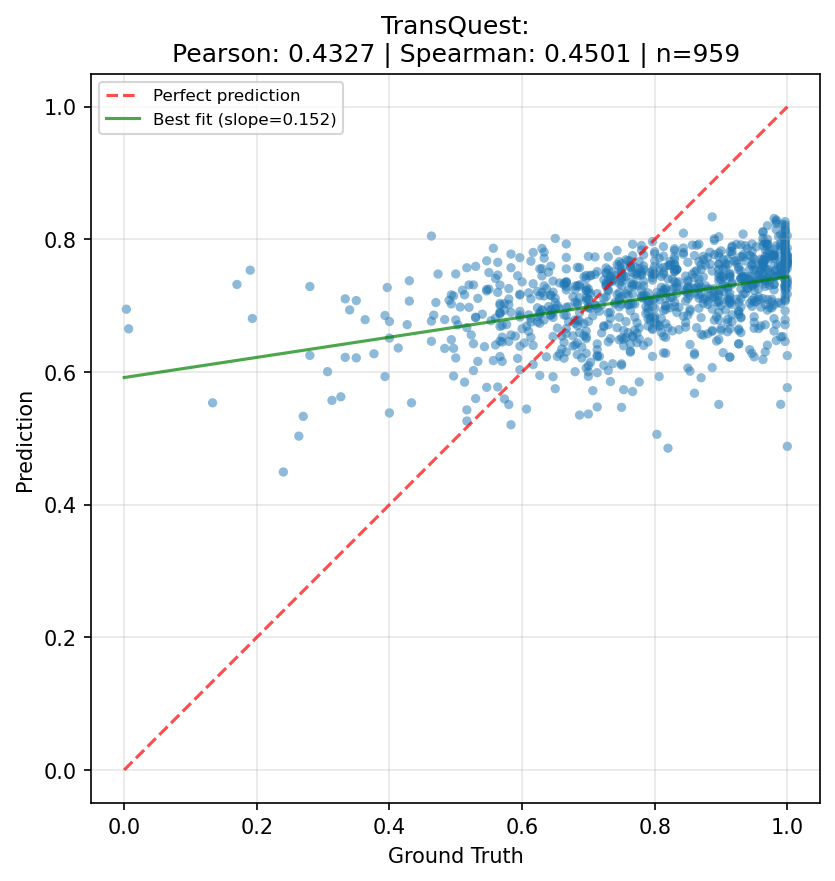}
    \caption{TransQuest}
  \end{subfigure}
  \hfill
  \begin{subfigure}{0.32\linewidth}
    \includegraphics[width=\linewidth]{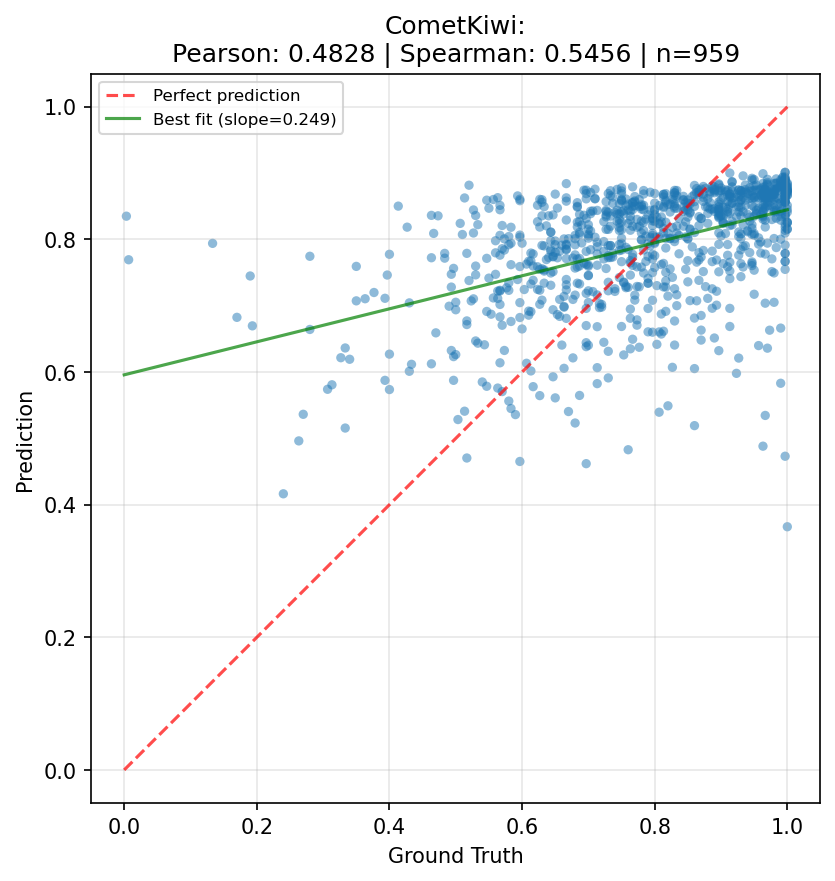}
    \caption{CometKiwi}
  \end{subfigure}
  \caption{Scatter plots of baseline model hypotheses on the full dataset (n=959).}
  \label{fig:baselineHyps}
\end{figure*}

We set the ground truth as the mean of the three annotators' scores,
and benchmark three baseline models:
 (1) ChatGPT \citep{openai_chatgpt} with two different prompts, (2) TransQuest, and (3) CometKiwi. 
 We report Pearson and Spearman correlation in Table \ref{tab:results} (``All''),
 and show scatter plots in Figure \ref{fig:baselineHyps}.

CometKiwi is the strongest of our baselines, with 0.4828 Pearson and 0.5456 Spearman.

We benchmark two versions of TransQuest: the ``en-any'' model performs best of the two,
with 0.4327 Pearson and 0.4501 Spearman compared to the ``multilingual'' (any-to-any)
version which lags behind at 0.3759 Pearson and 0.4303 Spearman.

ChatGPT-prompting produces results in between TransQuest and CometKiwi, with
0.4266 Pearson and 0.5018 Spearman.

Our ChatGPT prompts are (a) ``freestyle'': \texttt{``Score 0-100 by your own overall judgment of translation quality.''} and (b) ``guidelines'', where the prompt includes our full annotation guidelines (Appendix \ref{sec:chatGPTPrompts}).
The two ChatGPT prompts produce very similar results,
with almost identical score agreement (Figure \ref{fig:compareModelHypsGGGF}, Appendix \ref{sec:InterModelCorr}).
Our guidelines are based on the publicly available WMT QE task,
and we hypothesize that ChatGPT-freestyle may be implicitly following similar
guidelines memorized from pre-training on web data.

ChatGPT tends to favor repeating certain scores,
resulting in the horizontal bands in the plot in Figure \ref{fig:baselineHyps}(a) (left). By contrast, TransQuest and CometKiwi produce scores throughout the range.

Both TransQuest and CometKiwi never produce scores of 0.91 or higher, even for inputs with ground truth score above 0.91 (``excellent or perfect'' in our guidelines),
which accounts for 31\% of our dataset.
Thus, although both models are
strong baselines, there is room for improvement.

We also benchmark ensembling: averaging the models' predicted scores for each input sample. Following \textit{Occam's Razor} (preferring simpler solutions) we select the ``freestyle'' version of ChatGPT prompting for our Ensemble. As shown in Table \ref{tab:results}, \textbf{ensembling provides substantial improvement over CometKiwi alone}:
+6.4 percentage points Pearson (from 0.4828 to 0.5472)
and +5.6 percentage points Spearman (from 0.5456 to 0.6014).

\section{Fine-tuning Experiments}

\begin{table}[htbp]
\centering
\begin{tabular}{lr}
\hline
 Slice      &   Quantity \\
\hline
 train    &        300 \\
 validation &        100 \\
 test       &        559 \\
 \hline
 TOTAL       &        959 \\
\hline
\end{tabular}
\caption{Data Slices}
\label{tab:dataSlices}
\end{table}

We partition our dataset (Table \ref{tab:dataSlices}), then fine-tune TransQuest and CometKiwi for 50 steps (5 epochs) on \textit{train} (n=300) (following \citeauthor{jafari-harandi-etal-2024-epoque}) use \textit{validation} (n=100) for
checkpoint selection (by highest Pearson), and report on \textit{test} (n=559).

We use stratified sampling \citep{bishop2006pattern} for a similar distribution of scores across the slices. (See Figure \ref{fig:scoreDistEach} in Appendix \ref{sec:dataDist}.)
We perform this split five times with different random seeds and release the seeded splits with our dataset
for reproducibility.
We report the results on each seed in Tables \ref{tab:resEachSeedPearson} (Pearson) and \ref{tab:resEachSeedSpearman} (Spearman) in Appendix \ref{sec:resPerSeed}.

\subsection{Fine-tuning Methods}

We explore four fine-tuning methods: (1) Full fine-tuning [FullFT] where all of the model
parameters are free to change; (2) LoRA \citep{DBLP:journals/corr/abs-2106-09685},
which freezes the base model and learns low-rank updates to the attention and feed-forward layers;
(3) BitFit \citep{ben-zaken-etal-2022-bitfit} which fine-tunes only the bias terms
and the head classifier layer(s), corresponding to 0.2\% of the model's parameters; and (4) fine-tuning only the head classifier layer(s) [FTHead].
We use the same hyperparameters for all four methods, shown in Table \ref{tab:LoraHyps} (Appendix \ref{sec:FTSettings}), notably batch size 32 and learning rate 3e-5.

\subsection{Fine-tuning TransQuest}

We fine-tune from TransQuest DA
en-to-any,\footnote{HuggingFace: \href{https://huggingface.co/TransQuest/monotransquest-da-any_en}{TransQuest/monotransquest-da-en\_any}}
which is based on XLM-RoBERTa-large with
a standard classification head on top of the encoder.

\subsection{Fine-tuning CometKiwi}

CometKiwi\footnote{HuggingFace: \href{https://huggingface.co/Unbabel/wmt22-cometkiwi-da}{Unbabel/wmt22-cometkiwi-da}}
is based on InfoXLM-large \citep{chi-etal-2021-infoxlm} (XLM-RoBERTa-large architecture) with layer-wise attention and a specialized classification head, called the ``Estimator'' (three linear layers with tanh activation in between).
Crucially, we closely match
CometKiwi's architecture, including layerwise attention with dropout,
gamma scaling of layer mix,
and inputting the \textit{target first}, i.e. \texttt{[cls] target [sep] source [eos]}. We unfreeze the full Estimator head and layerwise attention for the parameter-efficient methods:
for LoRA, we configure this via \texttt{modules\_to\_save}. For BitFit and FTHead, we explicitly unfreeze them in our code.

\begin{table*}[h!t!]
\centering
\begin{tabular}{l|cc||cc}
\toprule
      & \multicolumn{2}{c||}{All (n=959)} & \multicolumn{2}{c}{Test (n=559)} \\
 Model                      & Pearson & Spearman &   Pearson &   Spearman  \\
\midrule
\multicolumn{5}{c}{Single Baseline Model} \\
\midrule
ChatGPT-freestyle [GPT-f] & 0.4266 & 0.5018 & 0.4136 ± 0.0157 & 0.5020 ± 0.0105 \\
ChatGPT-guidelines [GPT-g] & 0.4256 & 0.5074 & 0.4119 ± 0.0158 & 0.5087 ± 0.0097 \\
TransQuest-multilingual & 0.3759 & 0.4303 & 0.3608 ± 0.0325 & 0.4235 ± 0.0336 \\
\hline
TransQuest-en-any [TQ] & 0.4327 & 0.4501 & 0.4205 ± 0.0359 & 0.4537 ± 0.0375 \\
CometKiwi [CK] & \textbf{0.4828} & \textbf{0.5456} & \textbf{0.4495} ± 0.0179 & \textbf{0.5305} ± 0.0176 \\
\midrule
\multicolumn{5}{c}{Ensemble Baseline Models} \\
\midrule
Ensemble(GPT-f, TQ) & 0.5028 & 0.5622 & 0.4876 ± 0.0206 & 0.5608 ± 0.0130 \\
Ensemble(GPT-f, CK) & 0.5211 & 0.5929 & 0.4992 ± 0.0161 & 0.5798 ± 0.0101 \\
Ensemble(TQ, CK) & 0.5081 & 0.5459 & 0.4810 ± 0.0240 & 0.5390 ± 0.0274 \\ 
Ensemble(GPT-f, TQ, CK) & \textbf{0.5472} & \textbf{0.6014} & \textbf{0.5250} ± 0.0197 & \textbf{0.5926} ± 0.0146 \\
\midrule
\multicolumn{5}{c}{Fine-Tune TransQuest-en-any [TQ]} \\
\midrule
TQ+FullFT & -- & -- & 0.4287 ± 0.0230 & 0.4608 ± 0.0295 \\
TQ+LoRA & -- & -- & \textbf{0.4445} ± 0.0354 & \textbf{0.4828} ± 0.0390 \\
TQ+BitFit & -- & -- & 0.4424 ± 0.0344 & 0.4799 ± 0.0394 \\
TQ+FTHead & -- & -- & 0.4358 ± 0.0351 & 0.4718 ± 0.0368 \\
\midrule
\multicolumn{5}{c}{Fine-Tune CometKiwi [CK]} \\
\midrule
CK+FullFT & -- & -- & 0.4236 ± 0.0482 & 0.5034 ± 0.0325 \\
CK+LoRA & -- & -- & 0.4670 ± 0.0141 & \textbf{0.5554} ± 0.0138 \\
CK+BitFit & -- & -- & 0.4647 ± 0.0140 & 0.5551 ± 0.0136 \\
CK+FTHead & -- & -- & \textbf{0.4693} ± 0.0134 & 0.5449 ± 0.0137 \\
\bottomrule
\end{tabular}
\caption{Main results: Pearson and Spearman correlation of baseline, ensembled, and fine-tuned models against ground truth. The column ``All'' is on the full dataset. For ``Test'', we report the mean ± standard deviation across the 5 seeded splits. The best result in each column and each section is bolded.} 
\label{tab:results}
\end{table*}

\section{Fine-tuning Results}

As shown in Table \ref{tab:results} (column ``Test''),
full fine-tuning fails to improve TransQuest,
and degrades CometKiwi by 2-3 percentage points.
However, the parameter-efficient methods LoRA, BitFit, and FTHead
improve both TransQuest and CometKiwi by 2-3 percentage points.
LoRA and BitFit are generally on par, with FTHead slightly behind.

\begin{figure*}[t]
\centering
\setlength{\tabcolsep}{2pt}
\renewcommand{\arraystretch}{1.05}

\begin{tabular}{c c c}
\textbf{Original} & \textbf{After Fine-Tuning} & \textbf{Learning Curve} \\
\hline
\includegraphics[width=0.29\textwidth]
{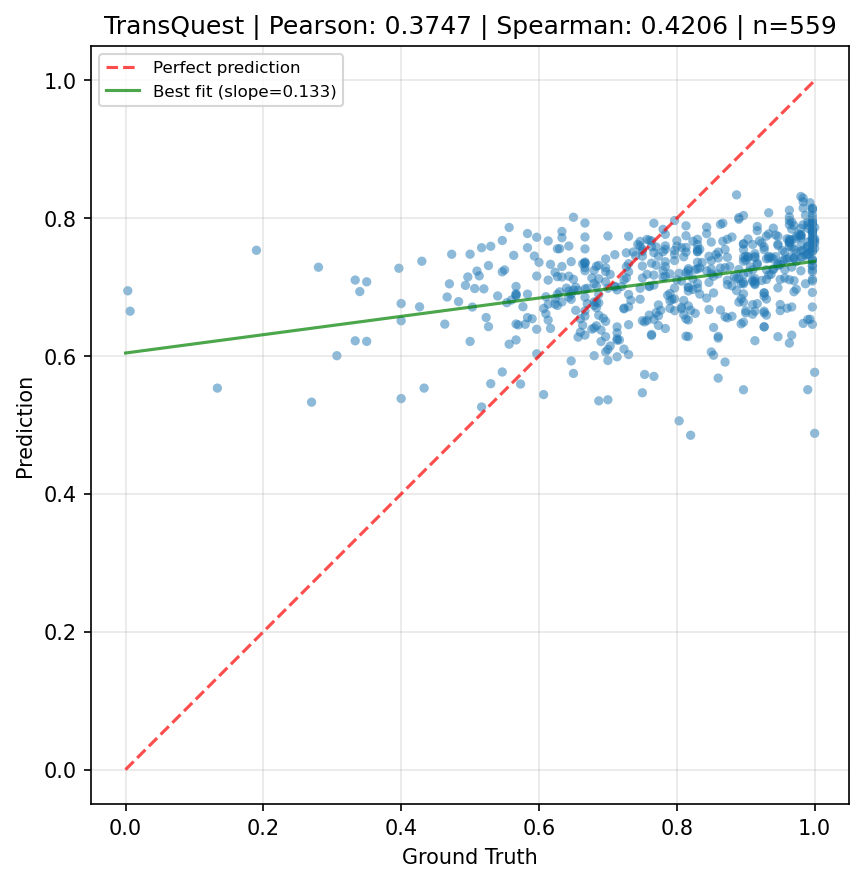} &
\includegraphics[width=0.29\textwidth]
{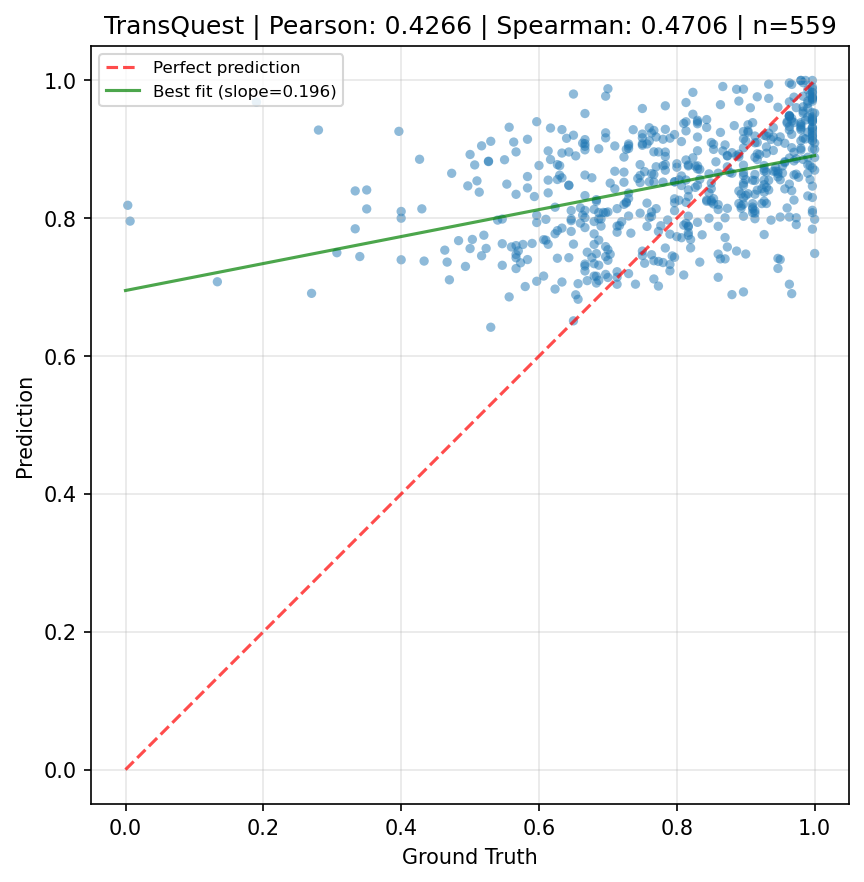} &
\includegraphics[width=0.29\textwidth]
{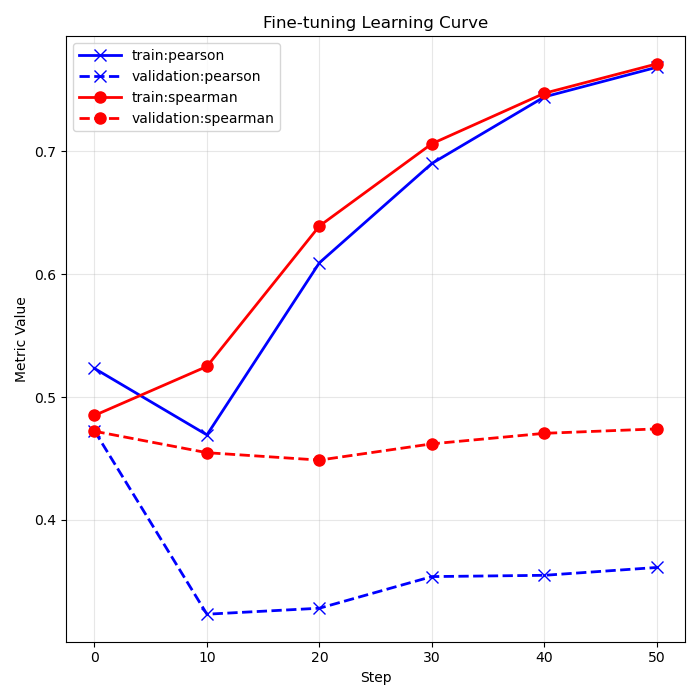} \\
\multicolumn{3}{c}{(a) TQ+FullFT} \\
\midrule

\includegraphics[width=0.29\textwidth]
{figures/scatter.tq_en_any.seed_0.test.png} &
\includegraphics[width=0.29\textwidth]
{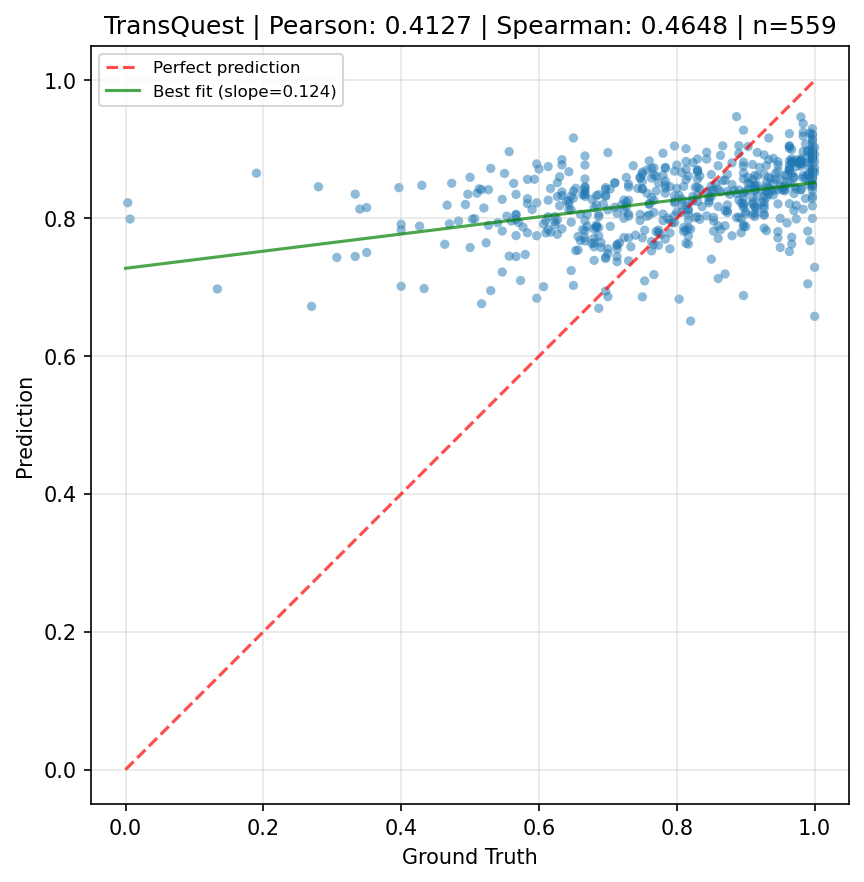} &
\includegraphics[width=0.29\textwidth]
{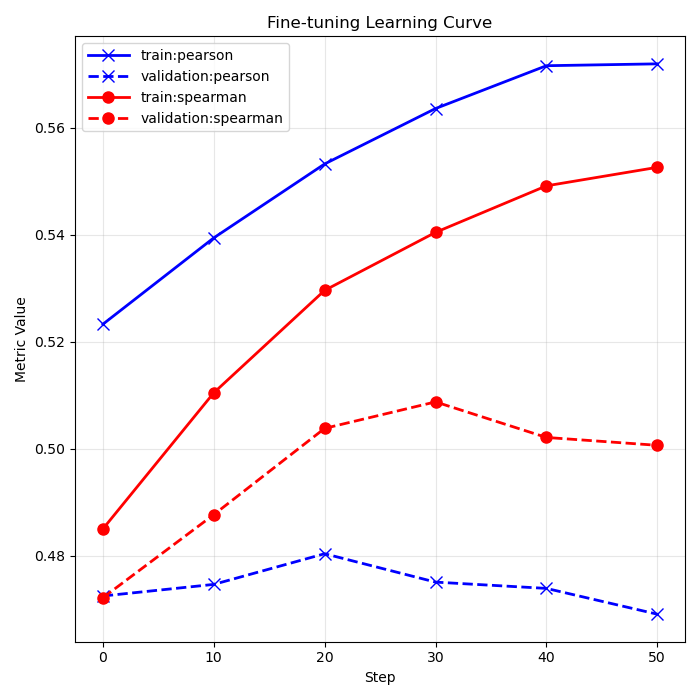} \\
\multicolumn{3}{c}{(b) TQ+LoRA} \\
\midrule

\includegraphics[width=0.29\textwidth]
{figures/scatter.tq_en_any.seed_0.test.png} &
\includegraphics[width=0.29\textwidth]
{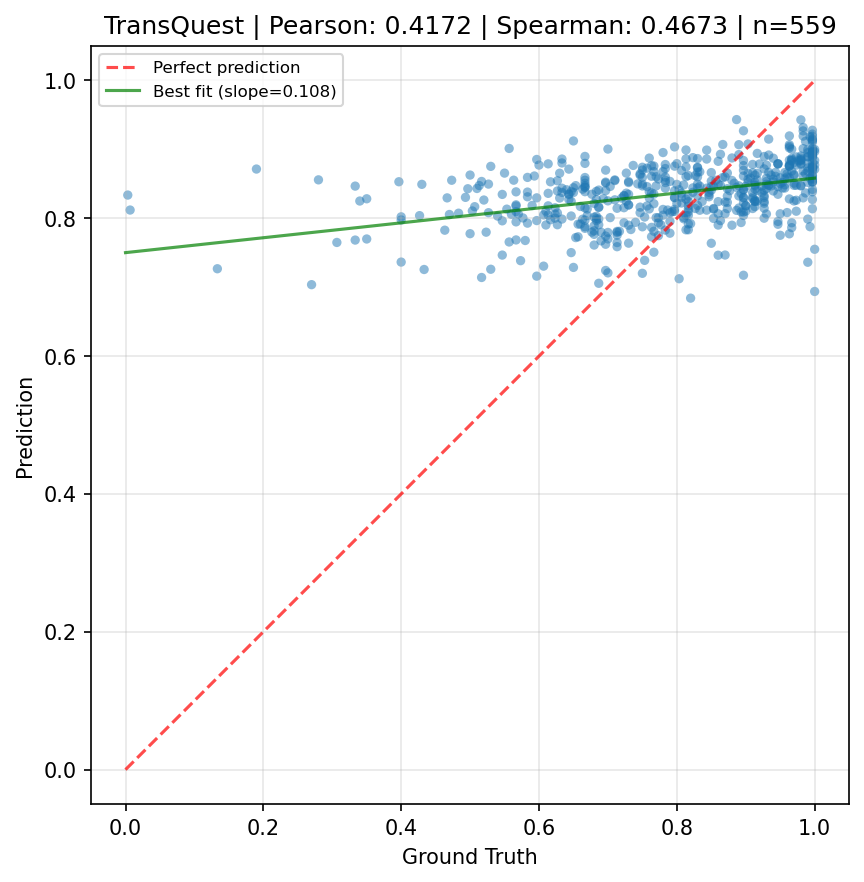} &
\includegraphics[width=0.29\textwidth]
{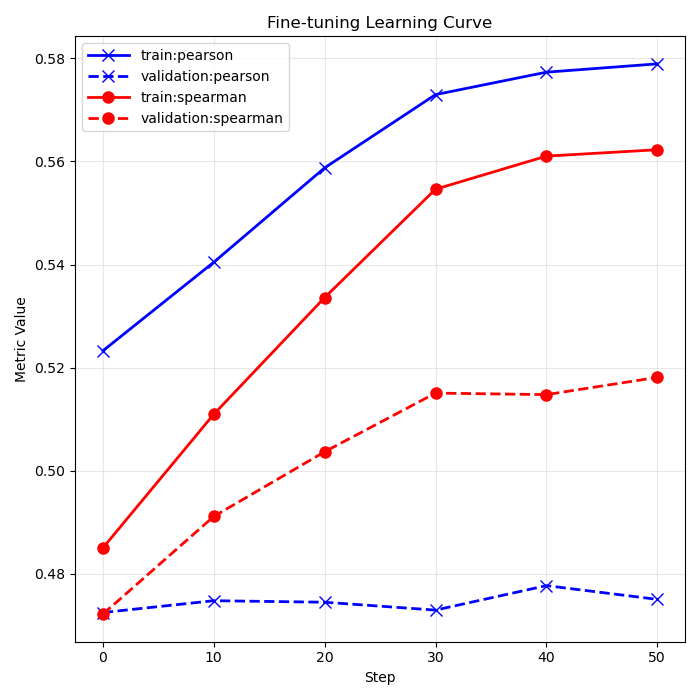} \\
\multicolumn{3}{c}{(c) TQ+BitFit} \\
\midrule

\includegraphics[width=0.29\textwidth]
{figures/scatter.tq_en_any.seed_0.test.png} &
\includegraphics[width=0.29\textwidth]
{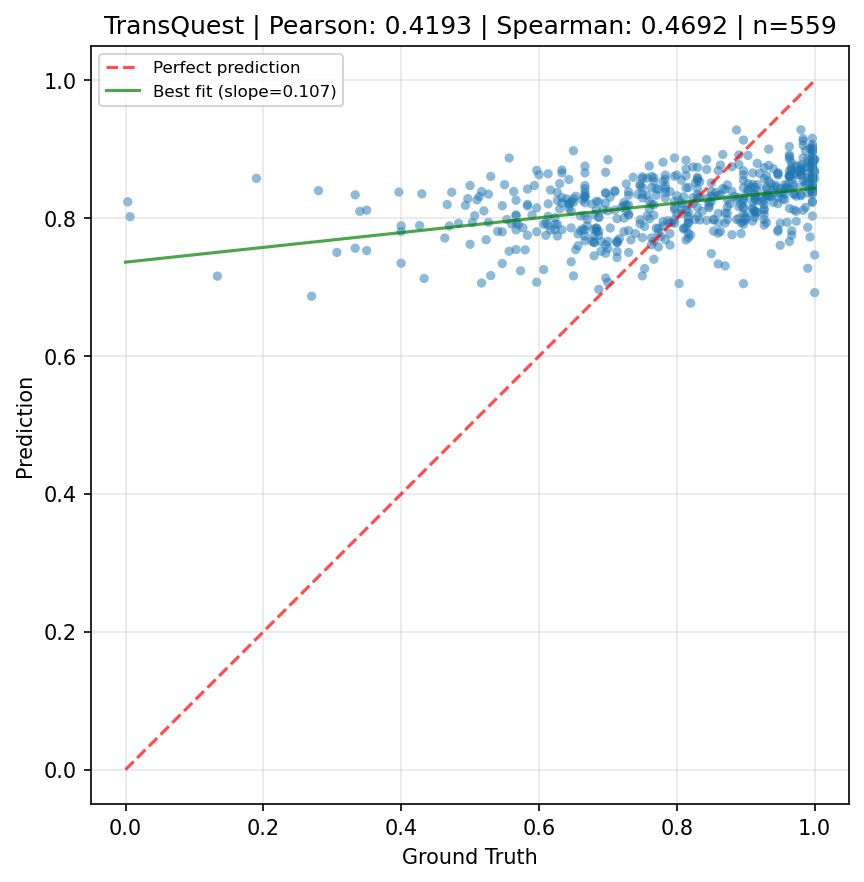} &
\includegraphics[width=0.29\textwidth]
{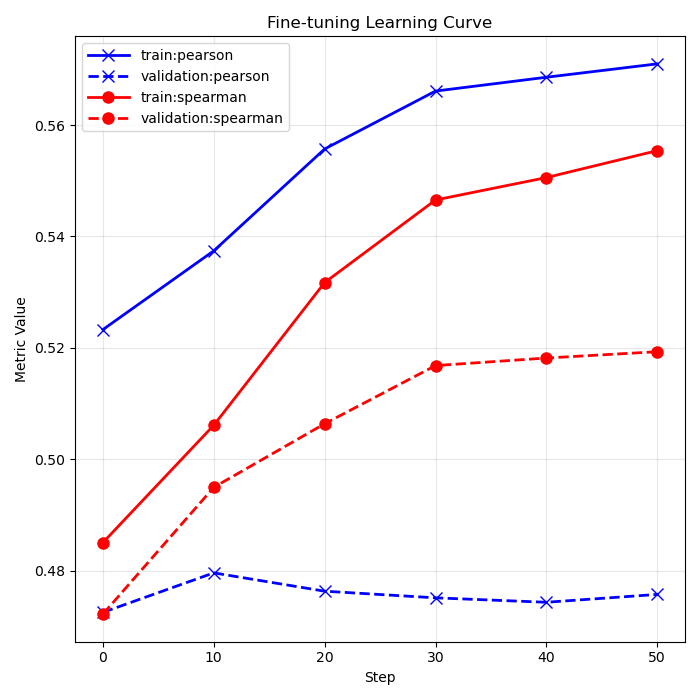} \\
\multicolumn{3}{c}{(d) TQ+FTHead} \\
\bottomrule

\end{tabular}
\caption{Visualization of effects of fine-tuning methods (seed 0). Each row represents the specified fine-tuning method. Left: baseline TransQuest scatter plot on test set (n=559), the same for each row. 
Center: best checkpoint scatter plot on the same test set.
Right: learning curve on train and validation during fine-tuning.
}
\label{fig:FTEffects}
\end{figure*}

For each method, we plot the score distribution
of TransQuest on the test set
before and after fine-tuning (for seed 0),
plus the learning curve (memorization of train data
vs. generalization to validation data), all in Figure \ref{fig:FTEffects}.
We further show the progression of the score distribution
on the validation data across epochs in Figure \ref{fig:FTEpochs} (Appendix \ref{sec:TQLearning}).

FullFT shows signs of overfitting and distribution collapse
(in Figure \ref{fig:FTEffects}(a), middle, the points
are more spread out than the baseline),
which could explain why FullFT does not improve on the test set.
However, the parameter-efficient methods show stable learning and modest
distribution shifts, consistent with their improvements on the test set.

\section{Conclusions}

We prepare, benchmark, and release \systemName{}, the first publicly available dataset for English-Hebrew Machine Translation Quality Estimation.
While ChatGPT prompting, TransQuest, and CometKiwi are
strong baselines, there is room to improve.
Light-weight fine-tuning (LoRA, BitFit, FTHead)
of TransQuest and CometKiwi on
300 samples from our data
improves Pearson and Spearman by 2-3 percentages points.
Future work can explore methods to improve
the prediction quality including synthetic data generation,
cross-lingual transfer, and calibration.
We hope that our work will unlock and inspire further research
on low-resource language systems, to benefit humanity.

\section*{Limitations}

MQM (Multidimensional Quality Metrics) are known to be superior
to DA scores \citep{GRAHAM_BALDWIN_MOFFAT_ZOBEL_2017}.
Due to budget constraints, we were limited to DA for this work,
which we still find useful.

We use the same set of hyperparameters for all fine-tuning methods,
which may not be optimal.

Our translations come from one system (Google Translate), which may limit generalization.

The modest dataset size (n=959) and high score distribution limit the number of low-score samples.
Future work can develop larger and more diverse datasets
to advance MTQE research for English-Hebrew and other low-resource language pairs.

\clearpage

\bibliography{custom}

\section*{Acknowledgments}

We thank the annotators for their diligent efforts in labeling the \systemName{} dataset
and the anonymous reviewers for providing feedback our work.
We use ChatGPT and Claude to help us write code, do literature review, and debug experiments.
We use HuggingFace \citep{wolf-etal-2020-transformers} (version 4.57.3) and PyTorch \citep{paszke2019pytorchimperativestylehighperformance} (version 2.9.1) for our experiments.

\appendix

\section{TransQuest Learning Dynamics}
\label{sec:TQLearning}

\begin{figure}[h]
\centering
\includegraphics[width=0.32\textwidth]{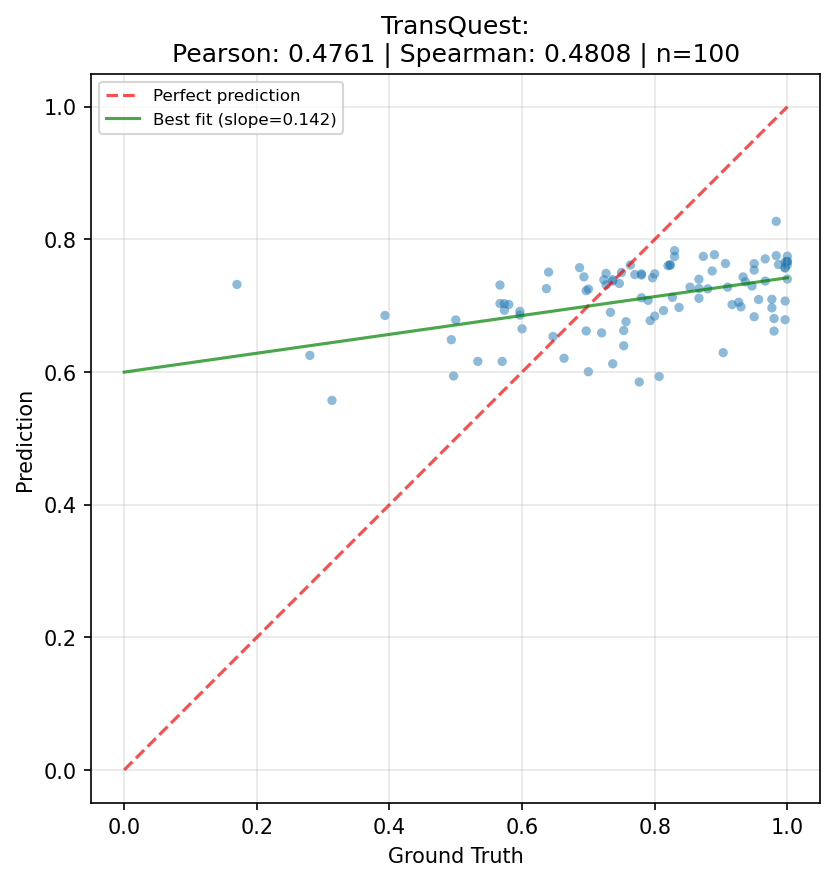}
\caption{Baseline TransQuest Scores on validation set (seed 0).
Starting point for fine-tuning plots in Figure \ref{fig:FTEpochs}.}
\label{fig:tqBaseDev}
\end{figure}

\begin{figure*}[h!t!]
\centering
\setlength{\tabcolsep}{2pt}
\renewcommand{\arraystretch}{1.05}

\begin{tabular}{c|c|c|c}
\textbf{TQ+FullFT} & \textbf{TQ+LoRA} & \textbf{TQ+BitFit} & \textbf{TQ+FTHead} \\

\includegraphics[width=0.24\textwidth]{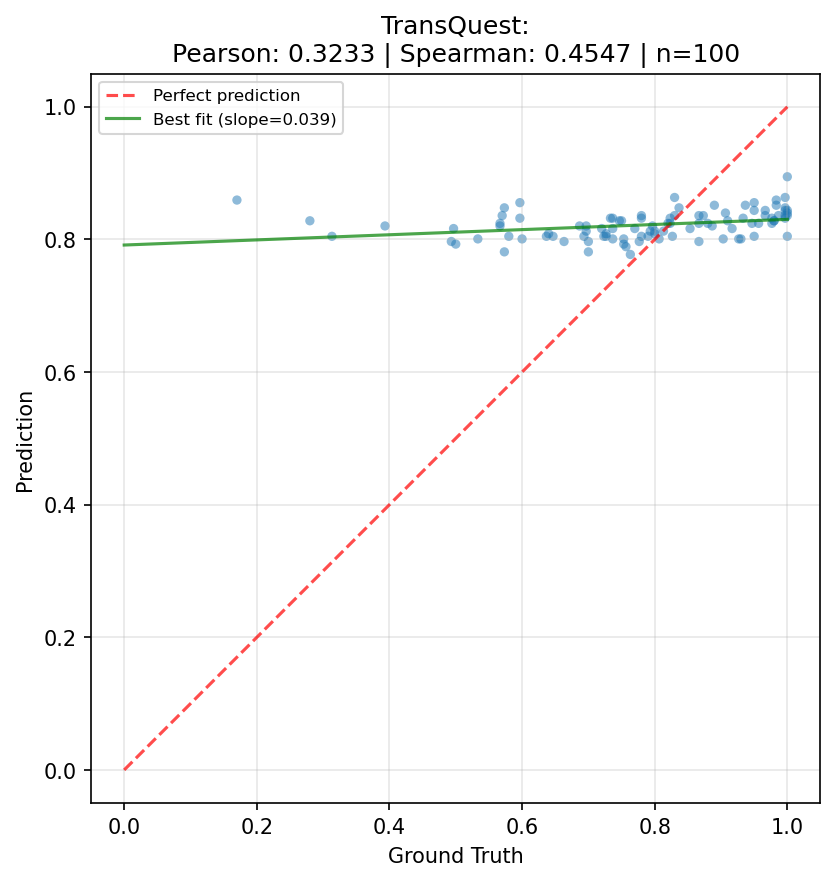} &
\includegraphics[width=0.24\textwidth]{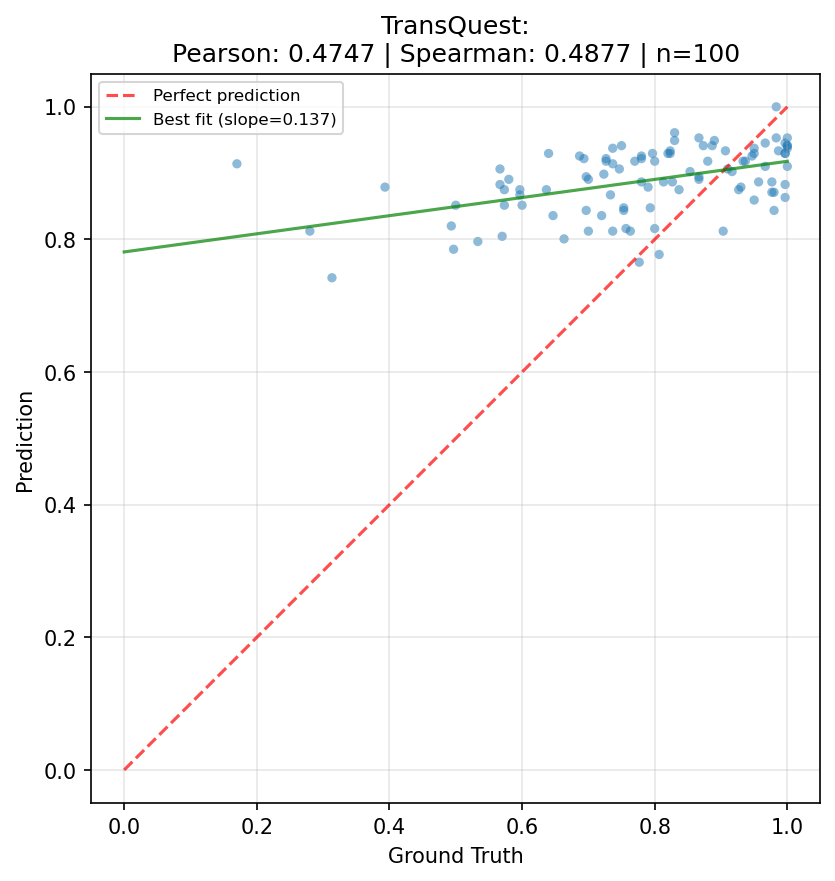} &
\includegraphics[width=0.24\textwidth]{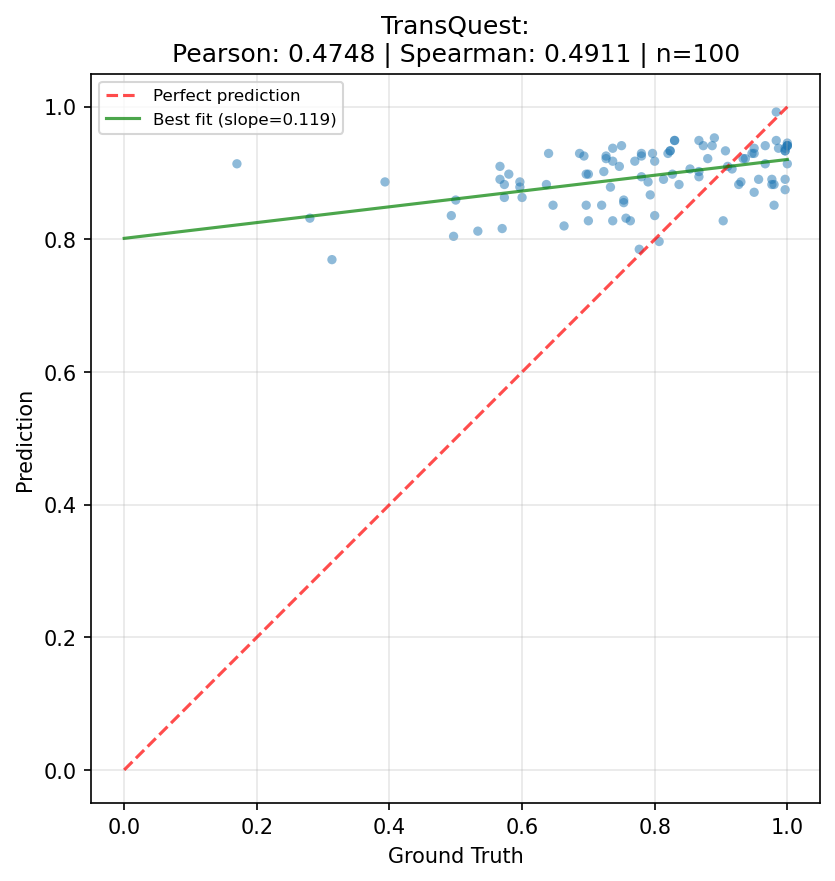} &
\includegraphics[width=0.24\textwidth]{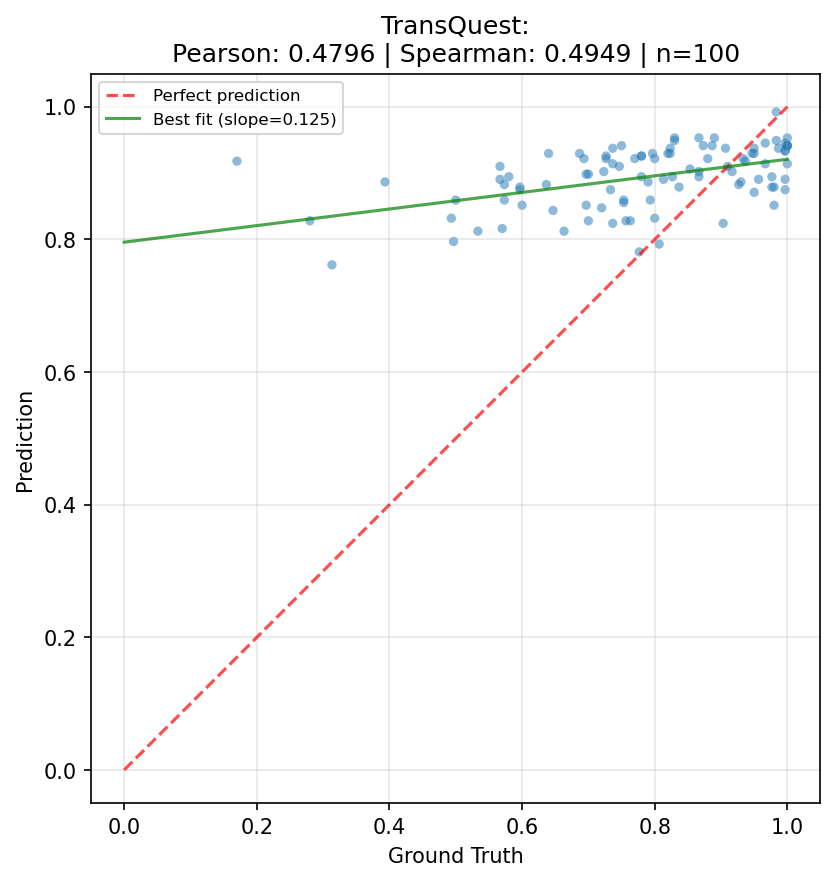} \\

\multicolumn{4}{c}{\textit{Epoch 1}} \\
\hline
\includegraphics[width=0.24\textwidth]{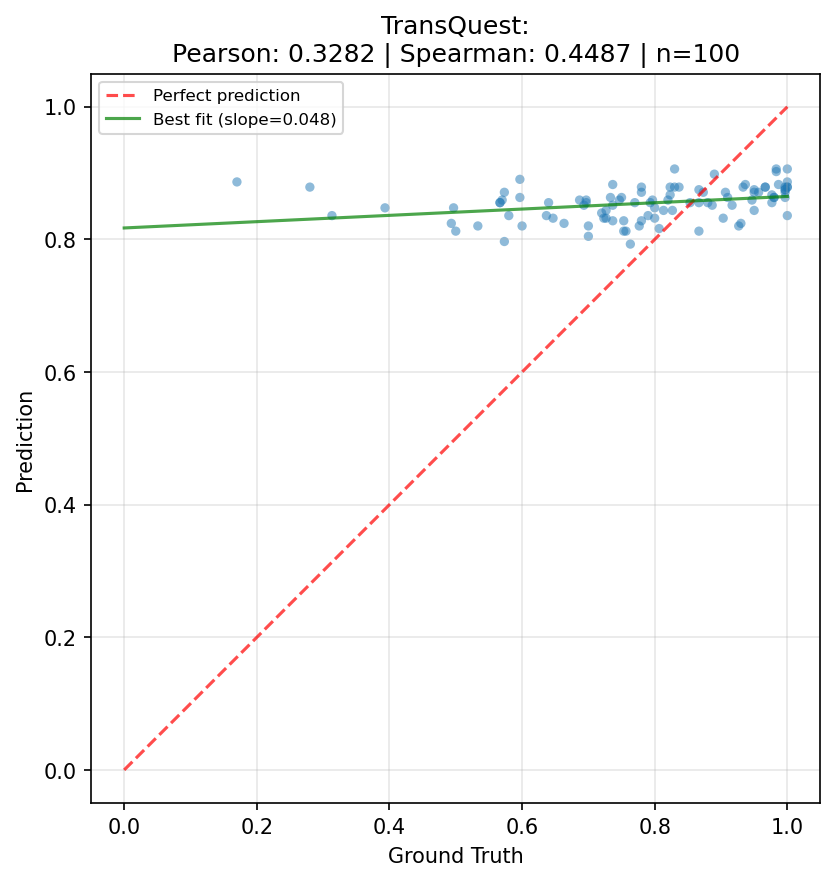} &
\includegraphics[width=0.24\textwidth]{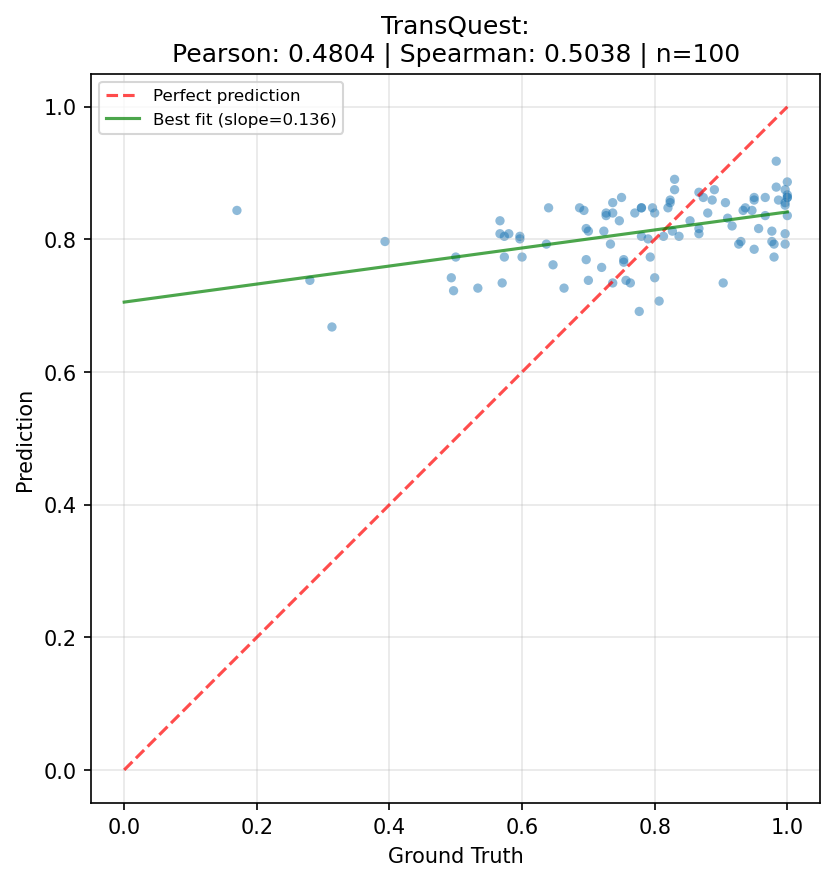} &
\includegraphics[width=0.24\textwidth]{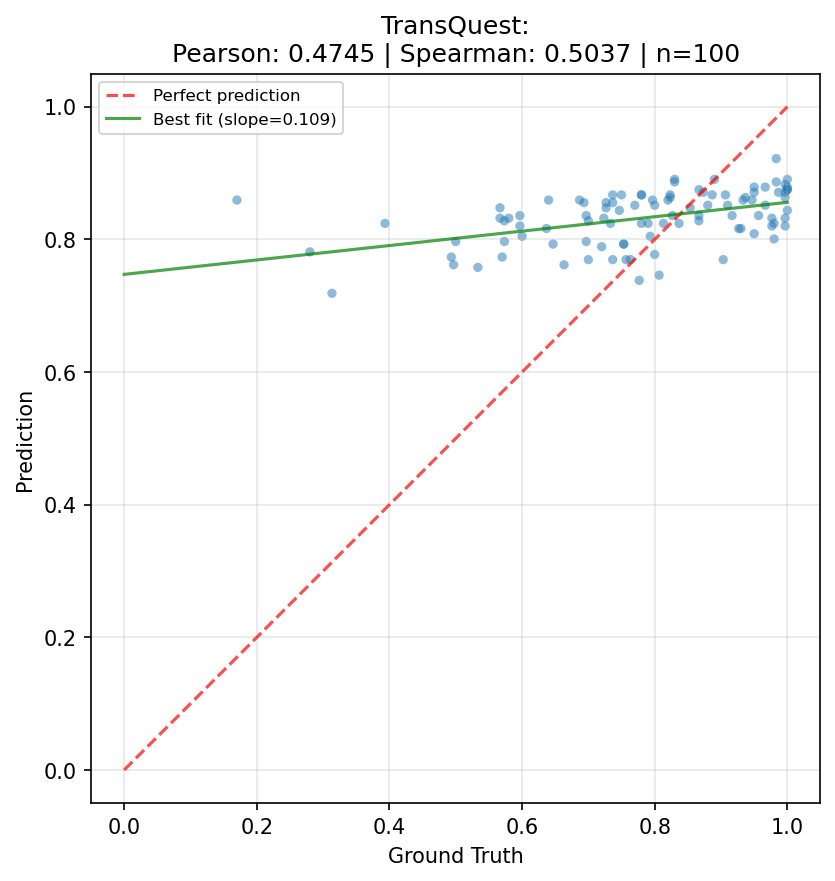} &
\includegraphics[width=0.24\textwidth]{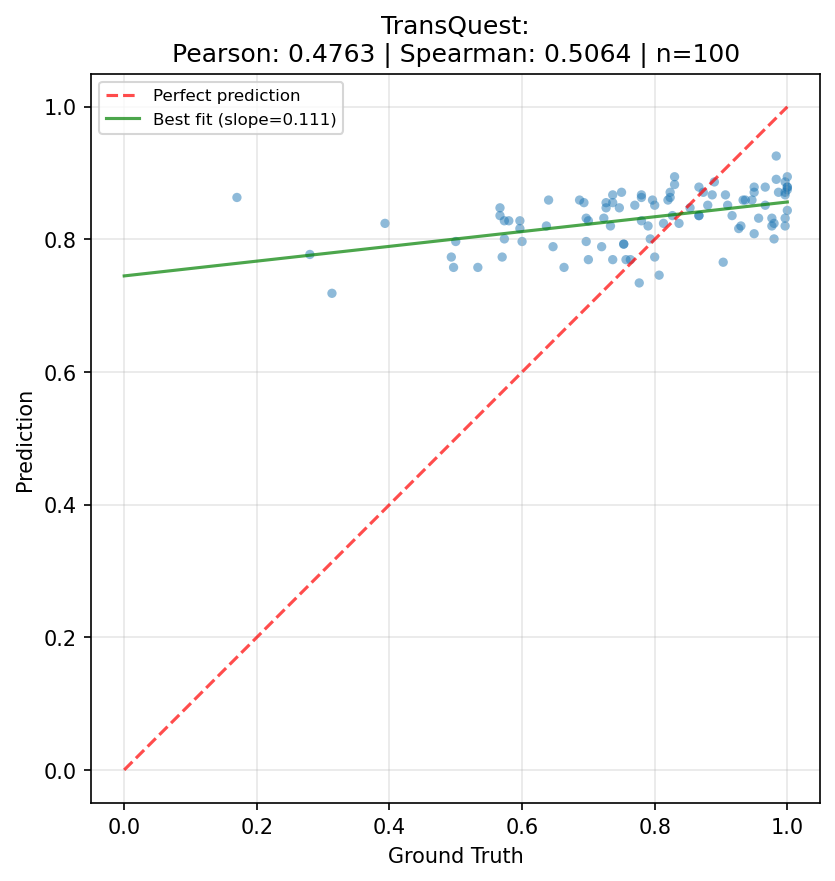} \\

\multicolumn{4}{c}{\textit{Epoch 2}} \\
\hline
\includegraphics[width=0.24\textwidth]{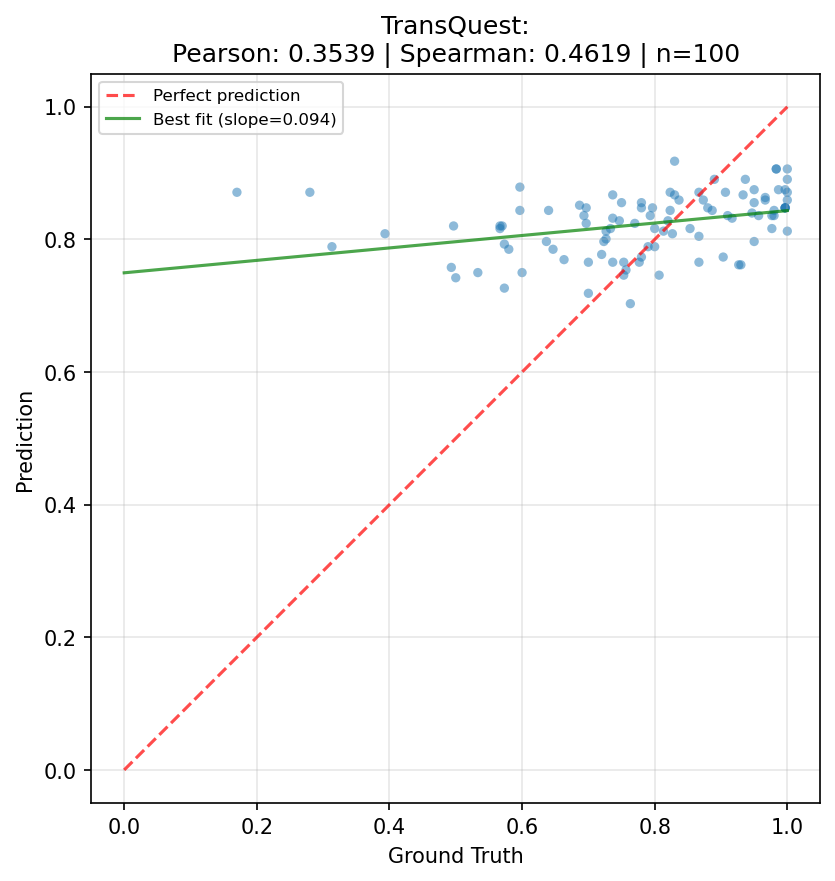} &
\includegraphics[width=0.24\textwidth]{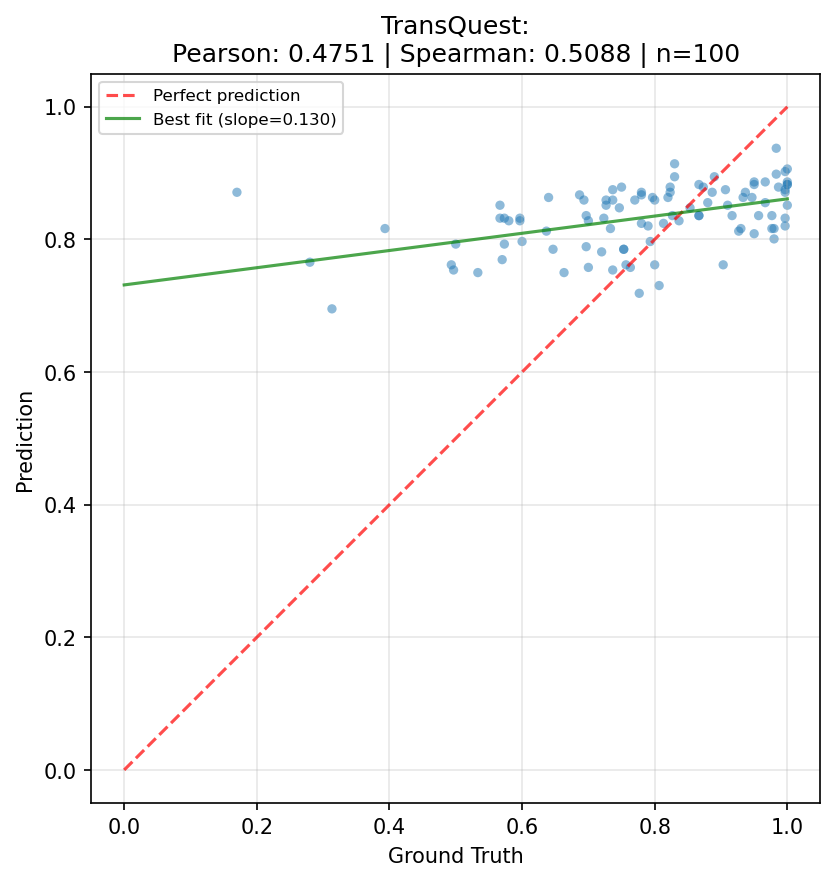} &
\includegraphics[width=0.24\textwidth]{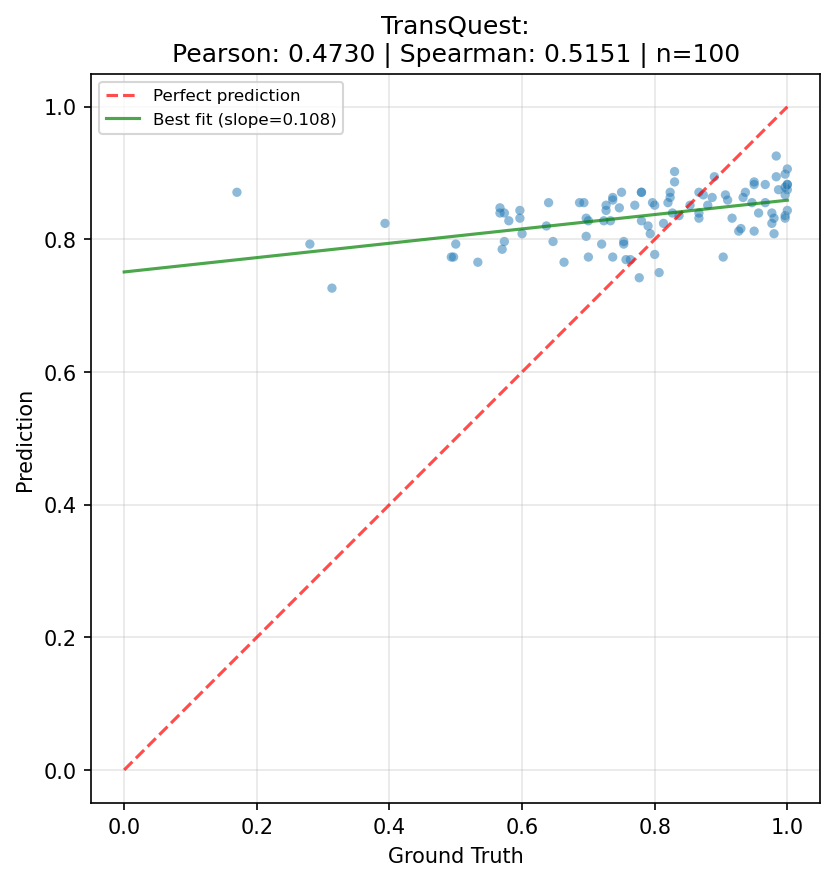} &
\includegraphics[width=0.24\textwidth]{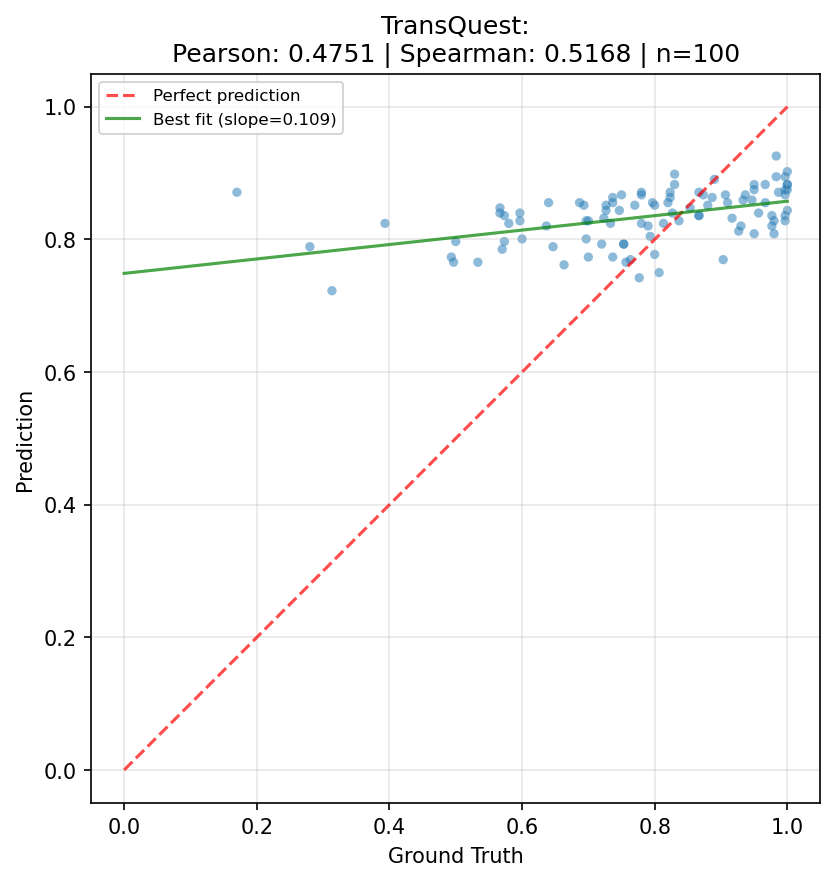} \\

\multicolumn{4}{c}{\textit{Epoch 3}} \\
\hline
\includegraphics[width=0.24\textwidth]{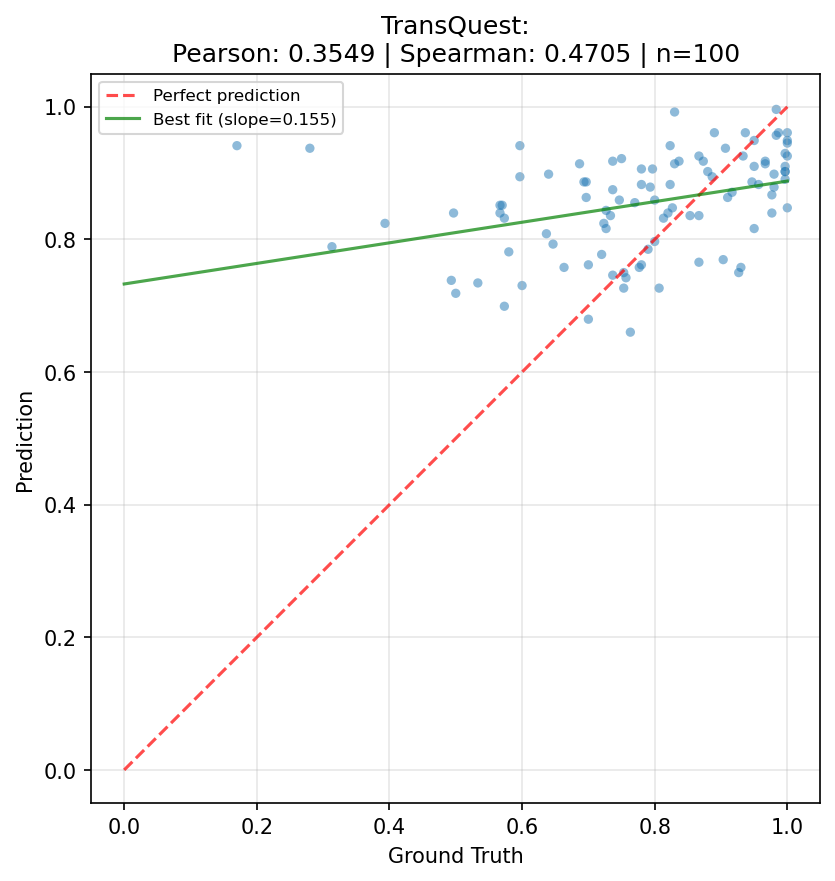} &
\includegraphics[width=0.24\textwidth]{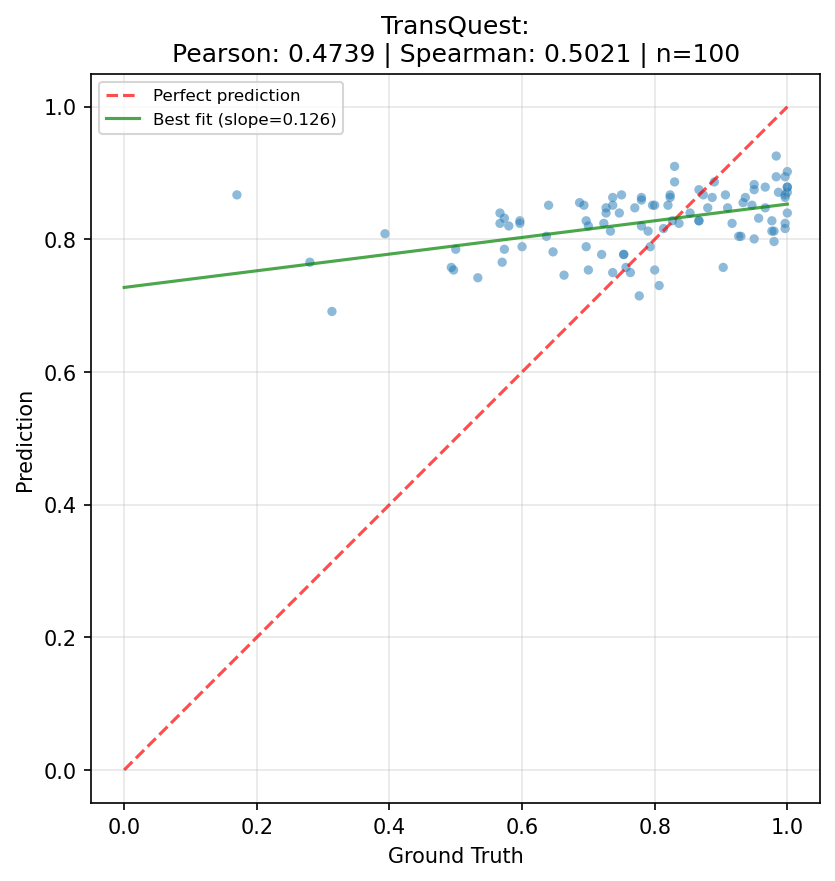} &
\includegraphics[width=0.24\textwidth]{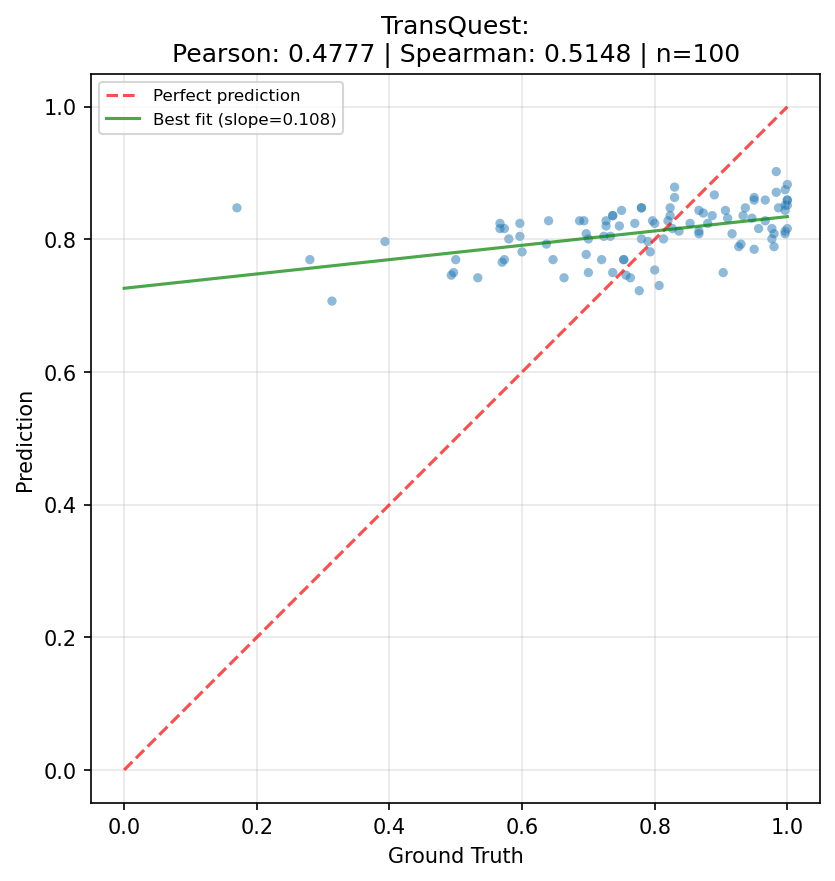} &
\includegraphics[width=0.24\textwidth]{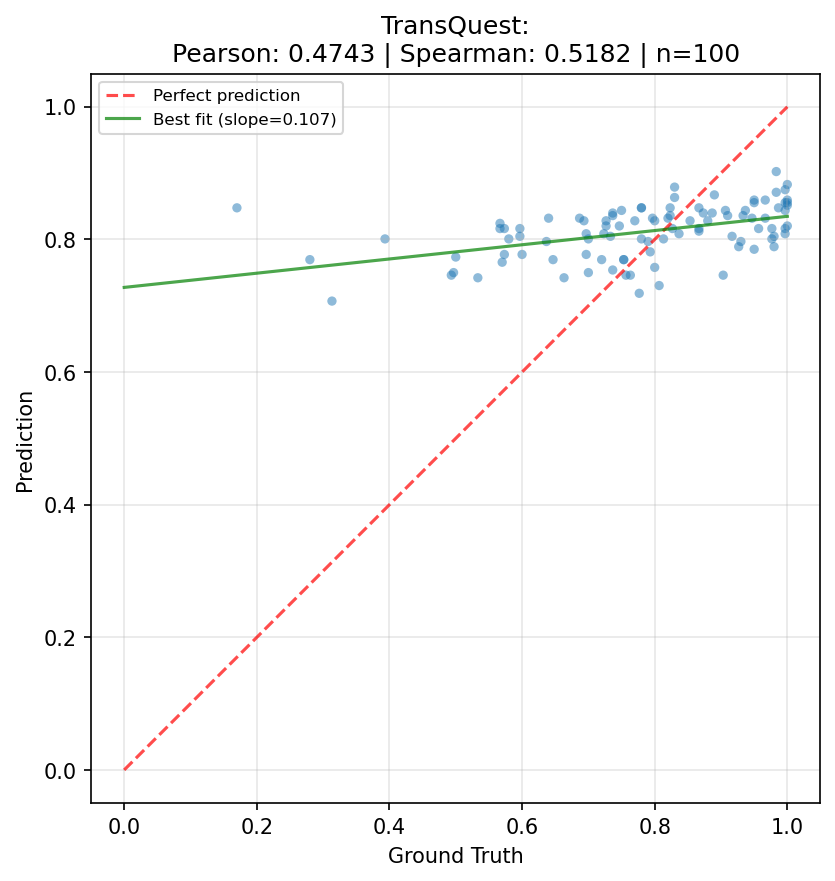} \\

\multicolumn{4}{c}{\textit{Epoch 4}} \\
\hline
\includegraphics[width=0.24\textwidth]{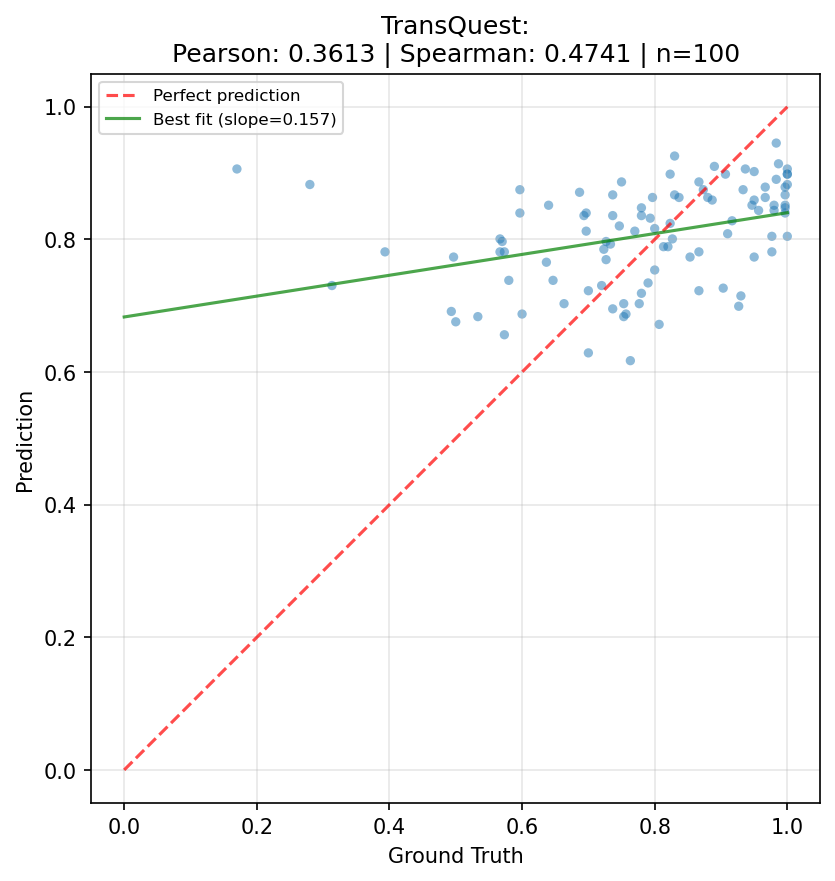} &
\includegraphics[width=0.24\textwidth]{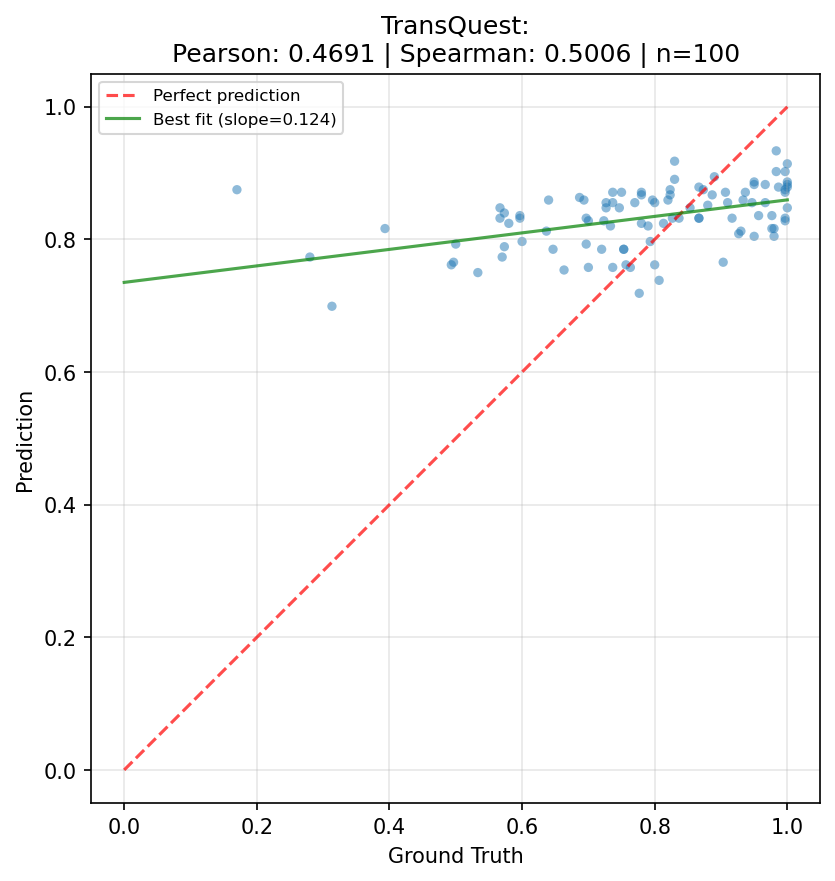} &
\includegraphics[width=0.24\textwidth]{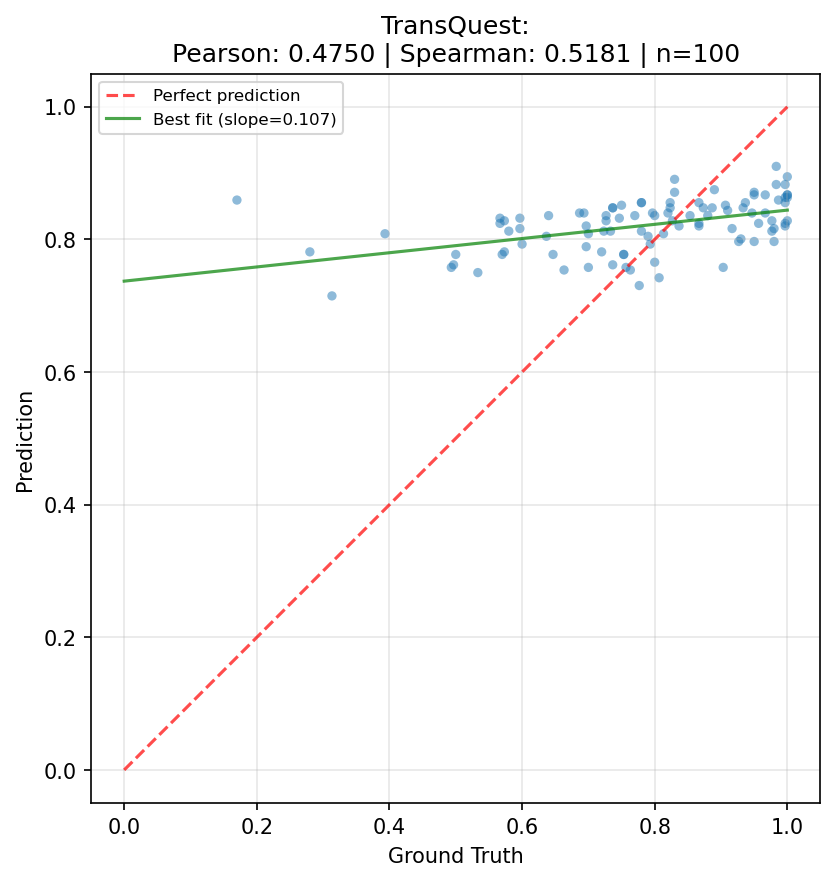} &
\includegraphics[width=0.24\textwidth]{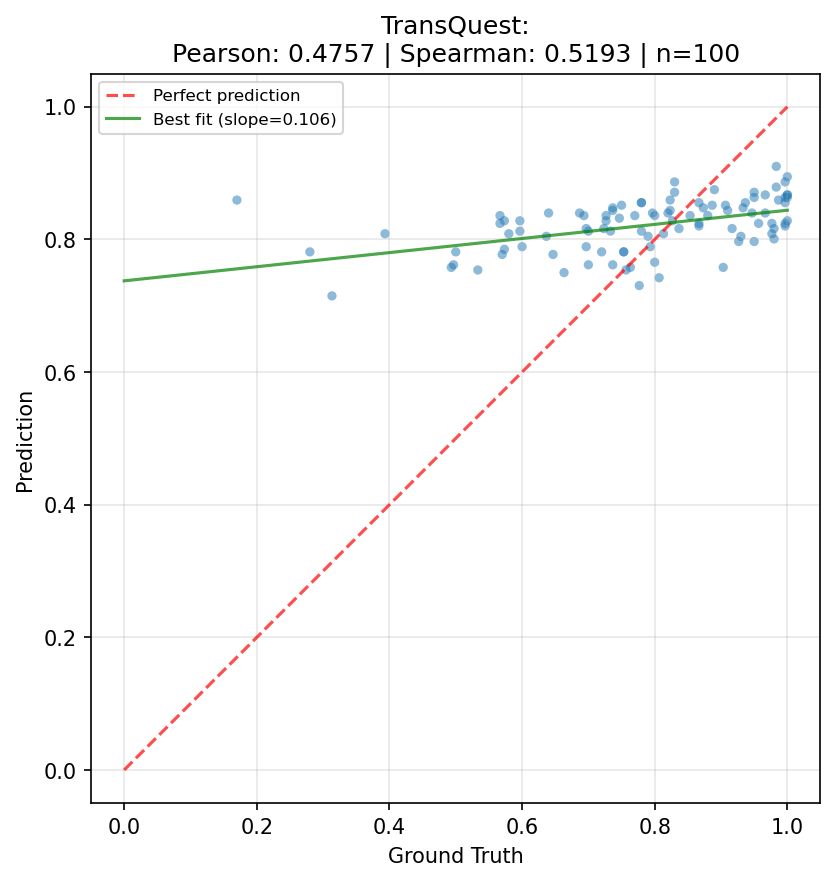} \\
\multicolumn{4}{c}{\textit{Epoch 5}} \\
\hline
\end{tabular}

\caption{
Scatter plots of predicted vs. truth score of fine-tuning methods across epochs.
The plots are on the validation set for seed 0 runs.
Rows correspond to epochs; columns correspond to fine-tuning strategies.
}
\label{fig:FTEpochs}
\end{figure*}

Figure \ref{fig:tqBaseDev} shows the baseline TransQuest
score distribution on the validation set (seed 0).
As we fine-tune this TransQuest model using
each of the four methods (FullFT, LoRA, BitFit, and FTHead),
we show in Figure \ref{fig:FTEpochs} how the score
distribution on the validation set evolves
over each epoch.

Notably, in FullFT, the distribution collapses after the
first epoch to be nearly flat, then disperses into
inconsistency in later epochs.
By contrast, the parameter-efficient methods train
smoothly with small and consistent distribution shifts.

\section{Inter-model Correlation}
\label{sec:InterModelCorr}

\begin{figure*}[h]
  \centering
  \begin{subfigure}{0.32\linewidth}
    \includegraphics[width=\linewidth]{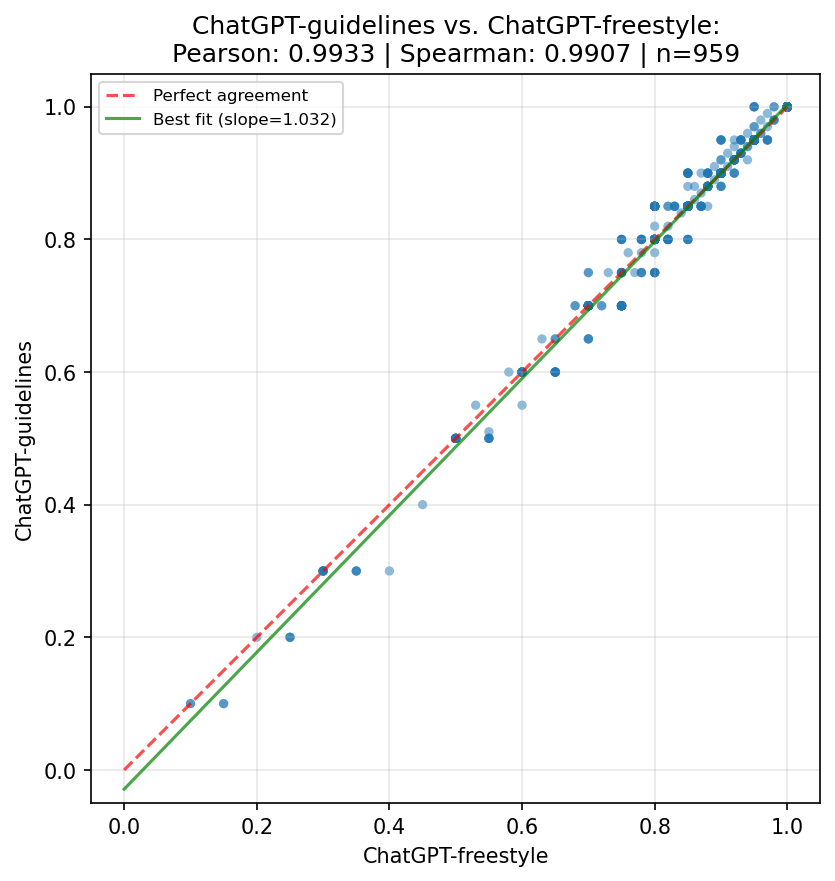}
    \caption{ChatGPT: guidelines vs. freestyle}
    \label{fig:compareModelHypsGGGF}
  \end{subfigure}
  \hfill
  \begin{subfigure}{0.32\linewidth}
    \includegraphics[width=\linewidth]{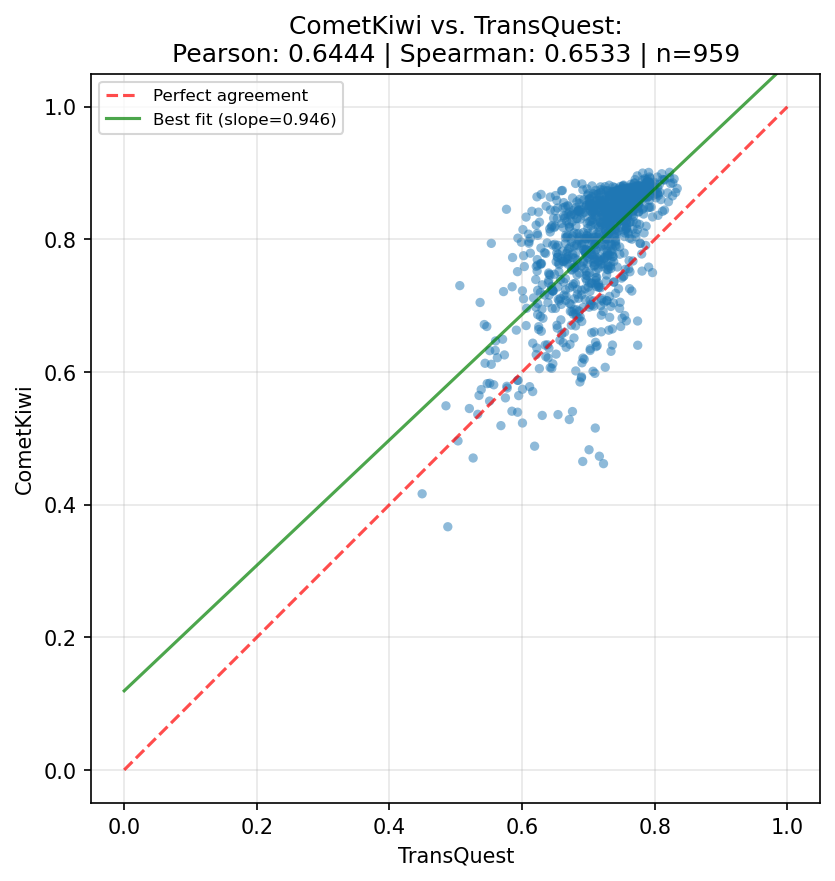}
    \caption{CometKiwi vs. TransQuest}
    \label{fig:compareModelHypsCKTQ}
  \end{subfigure}
  \begin{subfigure}{0.32\linewidth}
    \includegraphics[width=\linewidth]{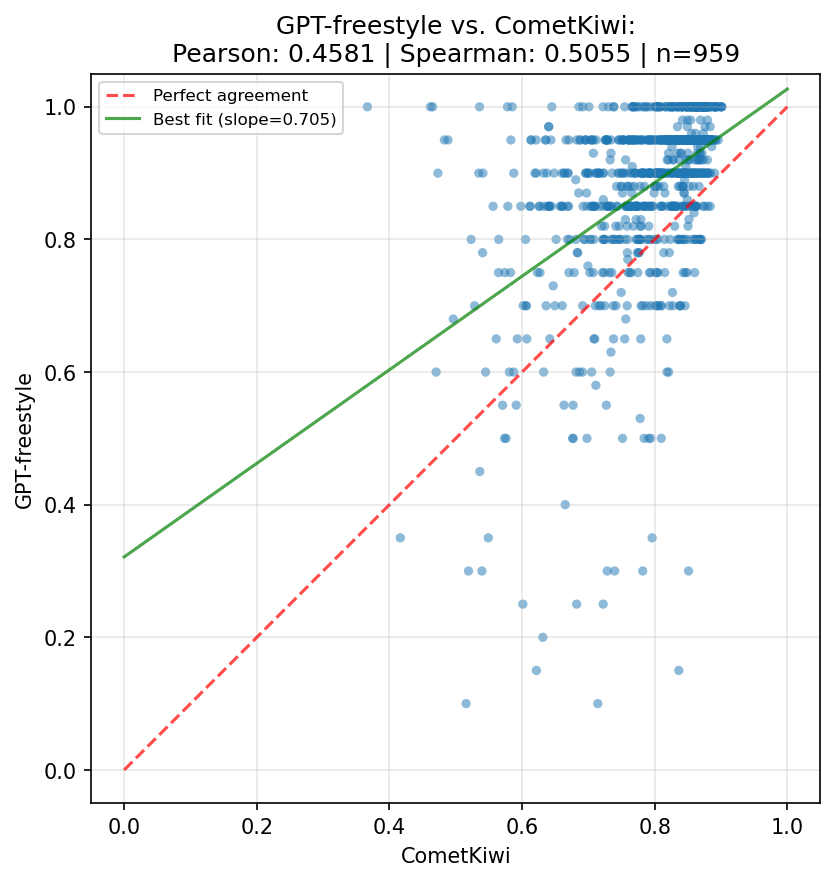}
    \caption{ChatGPT-freestyle vs. CometKiwi}
    \label{fig:compareModelHypsGFCK}
  \end{subfigure}
  \caption{Inter-Model Correlation of baseline models' hypotheses on the full dataset.}
  \label{fig:compareModelHyps}
\end{figure*}

To better understand the baseline models, we show the correlation between
their predictions in Figure \ref{fig:compareModelHyps}.
ChatGPT with and without the guidelines produces nearly identical scores,
as shown on the left (Figure \ref{fig:compareModelHypsGGGF}), with Pearson correlation 0.9933.
\mbox{TransQuest} and CometKiwi predictions differ somewhat more (Figure \ref{fig:compareModelHypsCKTQ}) with a Pearson
of only 0.6444. In particular, there is a cluster of samples
where both models predict a high range (0.8 to 0.9),
and they diverge more where one or the other predicts a lower score.
Finally, as shown on the right (Figure \ref{fig:compareModelHypsGFCK}),
ChatGPT and CometKiwi have moderate Pearson correlation
of 0.4581.

\section{Validation and Test Set Distribution}
\label{sec:dataDist}

\begin{figure*}[t]
  \centering
  \begin{subfigure}{0.32\linewidth}
    \includegraphics[width=\linewidth]{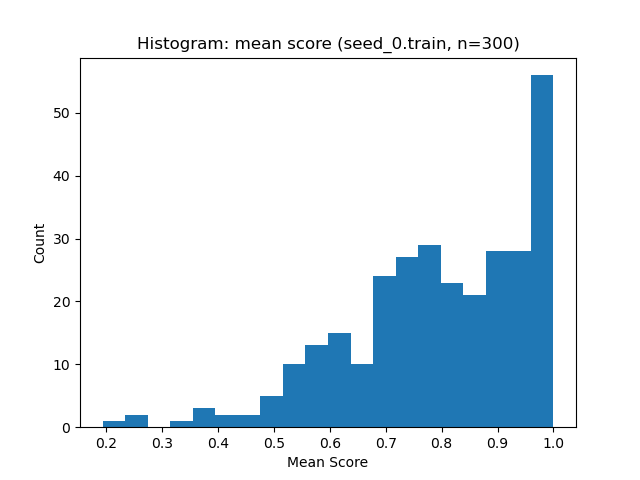}
    \caption{Train (n=300)}
  \end{subfigure}
  \hfill
  \begin{subfigure}{0.32\linewidth}
    \includegraphics[width=\linewidth]{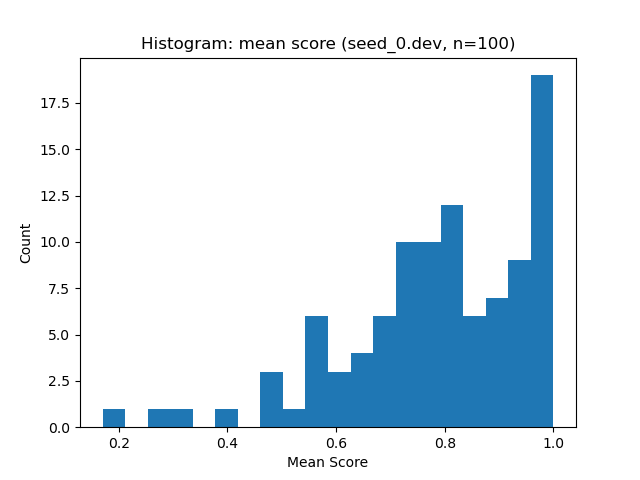}
    \caption{Validation (n=100)}
  \end{subfigure}
  \hfill
  \begin{subfigure}{0.32\linewidth}
    \includegraphics[width=\linewidth]{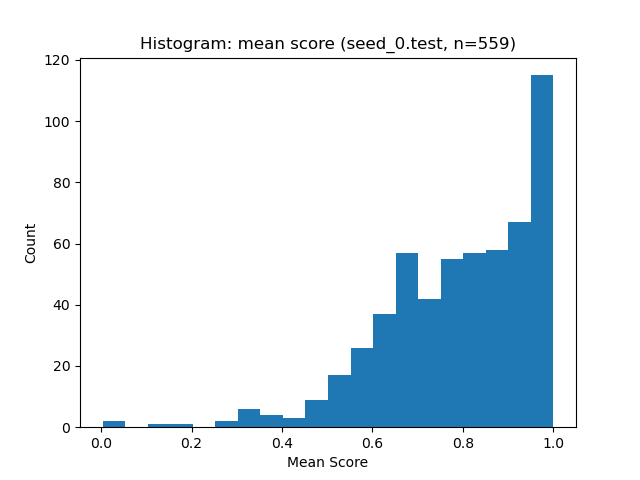}
    \caption{Test (n=559)}
  \end{subfigure}
  \caption{Score distribution on dataset slices (seed 0). Left: Train, Center: Validation, Right: Test.}
  \label{fig:scoreDistEach}
\end{figure*}

We show the score distribution for train (n=300), validation (n=100), and test (559)
for seed 0 in Figure \ref{fig:scoreDistEach}.
Comparing to Figure \ref{fig:dataStats}, we see that each slice has a similar distribution to the full dataset, with slight differences due to the small sample size.

\section{Influence of segment length}
\label{sec:segmentLen}

\begin{figure*}[h!]
  \centering
  \begin{subfigure}{0.32\linewidth}
    \includegraphics[width=\linewidth]{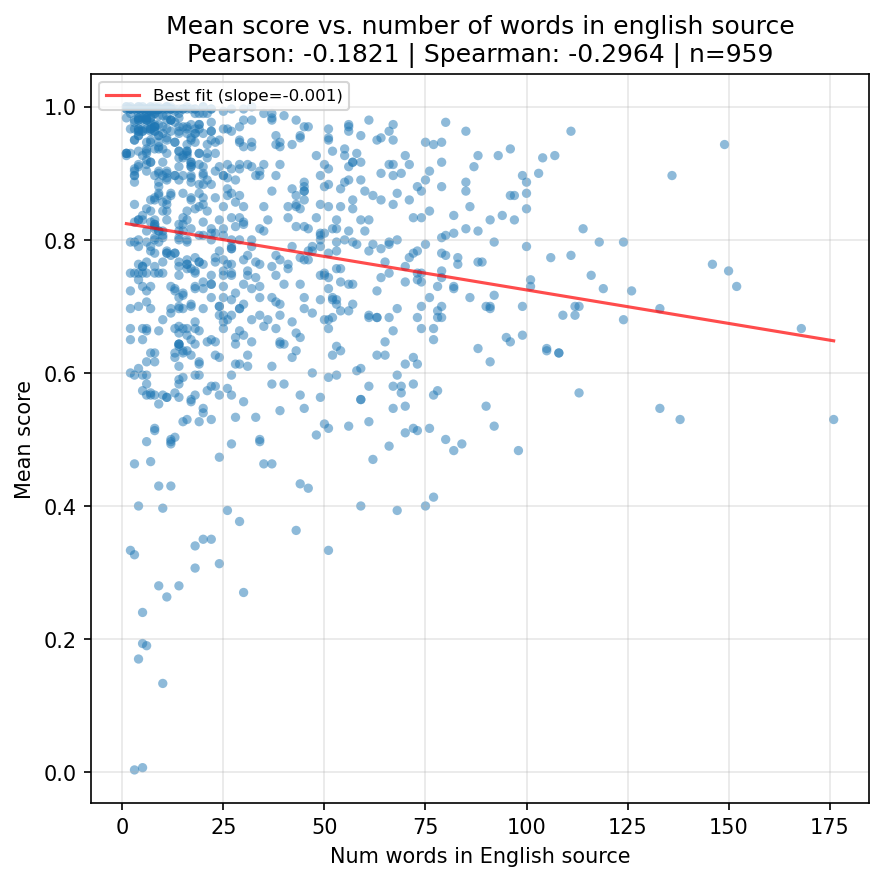}
    \caption{Annotator mean score vs. length.}
  \end{subfigure}
  \begin{subfigure}{0.36\linewidth}
    \includegraphics[width=\linewidth]{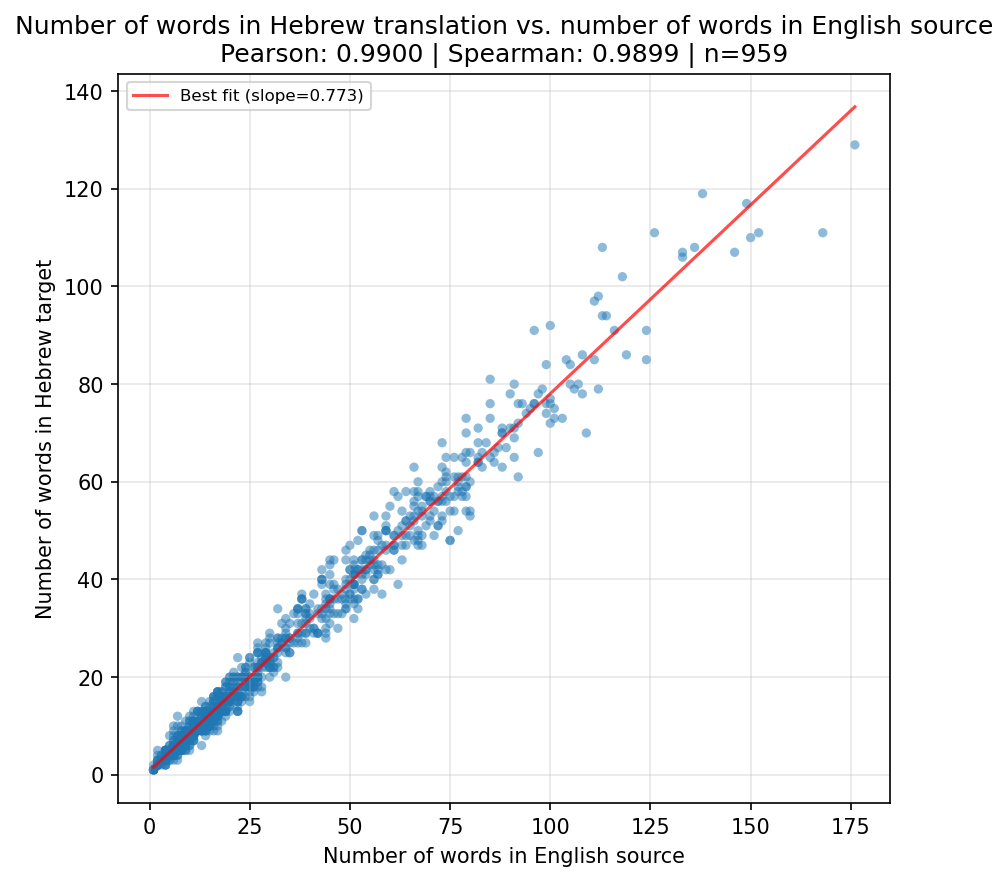}
    \caption{Hebrew length vs. English length.}
  \end{subfigure}
  \caption{Influence of source segment length on score distribution and translation length (full dataset).}
  \label{fig:lenPlots}
\end{figure*}

As shown in Figure \ref{fig:lenPlots}, the mean score label is slightly negatively correlated to
segment length (Pearson -18.21 and Spearman -0.2964
-- i.e., longer segments tend to have lower scores),
yet the plot shows also many short segments with low scores..
As expected from Google Translate which is a high quality
translation system, the Hebrew length is highly positively
correlated with the English length (Pearson 0.9900 and Spearman 0.9899). The Hebrew translation tends to have fewer words
than English (0.773 slope of best fit line), a result of
Hebrew's complex morphology which contributes to the difficulty
of our task.

\section{Fine-tuning Hyperparameters}
\label{sec:FTSettings}

We show our fine-tuning hyperparameters in Table \ref{tab:LoraHyps}.
We use the same hyperparameters for all four fine-tuning methods.

\begin{table}[htbp]
\centering
\begin{tabular}{l|c}
\toprule
 Parameter              & Value \\
\midrule
 Peak Learning Rate     & 3e-5        \\
 Learning Rate Schedule & \makecell{10\% warmup, \\ linear decay to 0} \\
 Batch Size             & 32 \\
 Max Num Steps          & 50                              \\
 Eval Every Steps       & 10                              \\
\hline
 LoRA r                 & 8                          \\
 LoRA alpha             & 16                       \\
 LoRA dropout           & 0.1                     \\
 LoRA modules           & query, key, value            \\
\bottomrule
\end{tabular}
\caption{Fine-tuning hyperparameters}
\label{tab:LoraHyps}
\end{table}

\section{Test Results for Each Seed}
\label{sec:resPerSeed}

We provide the test results for each seed in
Table \ref{tab:resEachSeedPearson} (Pearson)
and Table \ref{tab:resEachSeedSpearman} (Spearman).

\begin{table*}[b!p!]
\centering
\begin{tabular}{lrrrrr|rr}
\toprule
 model          &   seed=0 &   seed=1 &   seed=2 &   seed=3 &   seed=4 &   mean &   stdev \\
\midrule
 ChatGPT-freestyle [GPT-f] &   0.3968 &   0.4190 &   0.4318 &   0.3977 &   0.4229 & 0.4136 & ± 0.0157 \\
 ChatGPT-guidelines [GPT-g] &   0.3946 &   0.4159 &   0.4309 &   0.3967 &   0.4212 & 0.4119 & ± 0.0158 \\
 TransQuest-multilingual &   0.3093 &   0.3647 &   0.3561 &   0.3960 &   0.3780 & 0.3608 & ± 0.0325 \\
 TransQuest-en-any [TQ] &   0.3747 &   0.3926 &   0.4471 &   0.4291 &   0.4592 & 0.4205 & ± 0.0359 \\
 CometKiwi [CK] &   0.4300 &   0.4347 &   0.4492 &   0.4608 &   0.4730 & 0.4495 & ± 0.0179 \\
 \midrule
 Ensemble(GPT-f, TQ) &   0.4600 &   0.4837 &   0.5123 &   0.4791 &   0.5030 & 0.4876 & ± 0.0206 \\
 Ensemble(GPT-f, CK) &   0.4780 &   0.4960 &   0.5117 &   0.4919 &   0.5183 & 0.4992 & ± 0.0161 \\
 Ensemble(TQ, CK) &   0.4515 &   0.4611 &   0.4911 &   0.4917 &   0.5097 & 0.4810 & ± 0.0240 \\
 Ensemble(GPT-f, TQ, CK) &   0.4979 &   0.5173 &   0.5435 &   0.5213 &   0.5449 & 0.5250 & ± 0.0197 \\
\midrule
 TQ+FullFT &   0.4266 &   0.3944 &   0.4319 &   0.4317 &   0.4591 & 0.4287 & ± 0.0230 \\
 TQ+LoRA &   0.4127 &   0.4027 &   0.4688 &   0.4543 &   0.4842 & 0.4445 & ± 0.0354 \\
 TQ+BitFit &   0.4172 &   0.3978 &   0.4636 &   0.4514 &   0.4820 & 0.4424 & ± 0.0344 \\
 TQ+FTHead &   0.4193 &   0.3944 &   0.4630 &   0.4216 &   0.4806 & 0.4358 & ± 0.0351 \\
\midrule
 CK+FullFT &   0.3451 &   0.4347 &   0.4183 &   0.4470 &   0.4731 & 0.4236 & ± 0.0482 \\
 CK+LoRA &   0.4535 &   0.4597 &   0.4583 &   0.4771 &   0.4865 & 0.4670 & ± 0.0141 \\
 CK+BitFit &   0.4495 &   0.4573 &   0.4573 &   0.4790 &   0.4802 & 0.4647 & ± 0.0140 \\
 CK+FTHead &   0.4526 &   0.4617 &   0.4663 &   0.4816 &   0.4843 & 0.4693 & ± 0.0134 \\
\bottomrule
\end{tabular}
\caption{Test (n=559) results across random seeds for data split: Pearson Correlation against ground truth.}
\label{tab:resEachSeedPearson}
\end{table*}

\begin{table*}[b!p!]
\centering
\begin{tabular}{lrrrrr|rr}
\toprule
 model          &      seed=0 &      seed=1 &      seed=2 &      seed=3 &      seed=4 &   mean &   stdev \\
\midrule
 ChatGPT-freestyle [GPT-f] &   0.4877 &   0.5151 &   0.4961 &   0.5038 &   0.5072 & 0.5020 & ± 0.0105 \\
 ChatGPT-guidelines [GPT-g] &   0.4952 &   0.5219 &   0.5055 &   0.5116 &   0.5092 & 0.5087 & ± 0.0097 \\
 TransQuest-multilingual &   0.3754 &   0.4233 &   0.4090 &   0.4583 &   0.4516 & 0.4235 & ± 0.0336 \\
 TransQuest-en-any [TQ] &   0.4206 &   0.4063 &   0.4724 &   0.4812 &   0.4881 & 0.4537 & ± 0.0375 \\
 CometKiwi [CK] &   0.5416 &   0.5023 &   0.5307 &   0.5294 &   0.5483 & 0.5305 & ± 0.0176 \\
 \midrule
 Ensemble(GPT-f, TQ) &   0.5463 &   0.5484 &   0.5634 &   0.5711 &   0.5750 & 0.5608 & ± 0.0130 \\
 Ensemble(GPT-f, CK) &   0.5781 &   0.5696 &   0.5804 &   0.5744 &   0.5963 & 0.5798 & ± 0.0101 \\
 Ensemble(TQ, CK) &   0.5320 &   0.4966 &   0.5441 &   0.5524 &   0.5697 & 0.5390 & ± 0.0274 \\
 Ensemble(GPT-f, TQ, CK) &   0.5868 &   0.5724 &   0.5963 &   0.5950 &   0.6124 & 0.5926 & ± 0.0146 \\
\midrule
 TQ+FullFT &   0.4706 &   0.4119 &   0.4571 &   0.4770 &   0.4875 & 0.4608 & ± 0.0295 \\
 TQ+LoRA &   0.4648 &   0.4225 &   0.5024 &   0.5099 &   0.5146 & 0.4828 & ± 0.0390 \\
 TQ+BitFit &   0.4673 &   0.4172 &   0.4929 &   0.5091 &   0.5131 & 0.4799 & ± 0.0394 \\
 TQ+FTHead &   0.4692 &   0.4119 &   0.4918 &   0.4768 &   0.5093 & 0.4718 & ± 0.0368 \\
 \midrule
 CK+FullFT &   0.5197 &   0.5023 &   0.4649 &   0.4820 &   0.5483 & 0.5034 & ± 0.0325 \\
 CK+LoRA &   0.5691 &   0.5360 &   0.5512 &   0.5523 &   0.5684 & 0.5554 & ± 0.0138 \\
 CK+BitFit &   0.5686 &   0.5364 &   0.5492 &   0.5530 &   0.5681 & 0.5551 & ± 0.0136 \\
 CK+FTHead &   0.5571 &   0.5247 &   0.5425 &   0.5420 &   0.5582 & 0.5449 & ± 0.0137 \\
\bottomrule
\end{tabular}
\caption{Test (n=559) results across random seeds for data split: Spearman Correlation against ground truth.}
\label{tab:resEachSeedSpearman}
\end{table*}

\onecolumn  
\section{ChatGPT Prompt and Annotation Guidelines}
\label{sec:chatGPTPrompts}

\fbox{
    \begin{minipage}{\textwidth}
Direct Assessment Annotation Instructions (Official Standard)
\\ \\
Task\\
You will evaluate the quality of a translation by reading the original source sentence and the machine-translated output.\\
Your goal is to judge how accurately and fluently the translation conveys the meaning of the source.\\ \\

What to do\\
For each item:\\
1. Read the source sentence.\\
2. Read the translation.\\
3. Rate the translation on a scale from 0 to 100, where:\\
   - 0 = completely incorrect or nonsensical\\
   - 100 = perfect translation\\
4. Always give a score (don’t leave empty)\\ \\

Detailed Guidelines\\
- Consider adequacy (meaning correctness) and fluency (natural and grammatical language).\\
- Your score should reflect both aspects.\\
- If the translation is generally correct but contains some errors, deduct points proportionally.\\
- If the meaning is wrong—even if the text is fluent—the score must be low.\\
- URLs, Emails, hashtags, etc. – should remain untouched by the translation machine.\\
- Do not attempt to correct the translation.\\
- Do not compare with other translations.\\
- Do not use external tools or references.\\ \\

Scoring guide (use the entire scale)\\
0–10   Wrong meaning or incomprehensible\\
11–29  Some words match but overall meaning incorrect\\
30–50  Understandable but contains major errors\\
51–69  Mostly correct meaning; noticeable issues\\
70–90  Good translations with minor issues\\
91–100 Excellent or perfect\\ \\

Quality Control\\
You will occasionally see:\\
- High-quality human references\\
- Synthetic translations containing obvious errors\\
Please annotate these items normally. They help us track annotation quality.\\ \\

Important\\
- Work carefully and consistently.\\
- If unsure, choose the closest score—avoid always picking the same region.\\
- You may take breaks.\\
- All your responses are confidential.
  \end{minipage}
}
\end{document}